\documentclass{svjour3}

\usepackage[utf8]{inputenc}
\usepackage[T1]{fontenc}
\usepackage{lmodern}
\usepackage{fancyvrb}
\usepackage{graphicx}
\usepackage{algorithm}
\usepackage{algorithmic}
\usepackage{epsfig}
\usepackage{amsmath}
\usepackage{amsfonts}
\usepackage{diagbox}
\usepackage{wrapfig}
\usepackage{subcaption}
\usepackage{pdflscape}
\usepackage{afterpage}
\usepackage[table]{xcolor}
\usepackage[misc,geometry]{ifsym}

\journalname{Machine learning}
\begin{document}

\title{Explanatory machine learning for sequential human teaching}
\author{Lun Ai \and Johannes Langer \and Stephen H. Muggleton \and Ute Schmid}
\institute{Lun Ai \Letter \at
           Department of Computing, Imperial College London, London, UK\\
           \email{lun.ai15@imperial.ac.uk}
           \and
           Johannes Langer\at
           University of Bamberg, Bamberg, Germany\\
           \email{johannes-miran.langer@stud.uni-bamberg.de}
           \and
           Stephen H. Muggleton \at
           Department of Computing, Imperial College London, London, UK\\
           \email{s.muggleton@imperial.ac.uk}
           \and
           Ute Schmid\at
           Cognitive Systems Group, University of Bamberg, Bamberg, Germany\\
           \email{ute.schmid@uni-bamberg.de}
}
\date{}

\maketitle
\raggedbottom
\begin{abstract}
\vspace{-60pt}
The topic of comprehensibility of machine-learned theories has recently drawn increasing attention. Inductive Logic Programming (ILP) uses logic programming to derive logic theories from small data based on abduction and induction techniques. Learned theories are represented in the form of rules as declarative descriptions of obtained knowledge. In earlier work, the authors provided the first evidence of a measurable increase in human comprehension based on machine-learned logic rules for simple classification tasks. In a later study, it was found that the presentation of machine-learned explanations to humans can produce both beneficial and harmful effects in the context of game learning. We continue our investigation of comprehensibility by examining the effects of the ordering of concept presentations on human comprehension. In this work, we examine the explanatory effects of curriculum order and the presence of machine-learned explanations for sequential problem-solving. We show that 1) there exist tasks A and B such that learning A before learning B results in better comprehension for humans in comparison to learning B before learning A and 2) there exist tasks A and B such that the presence of explanations when learning A contributes to improved human comprehension when subsequently learning B. We propose a framework for the effects of sequential teaching on comprehension based on an existing definition of comprehensibility and provide evidence for support from data collected in human trials. Our empirical study involves curricula that teach novices the merge sort algorithm. Our results show that sequential teaching of concepts with increasing complexity a) has a beneficial effect on human comprehension and b) leads to human re-discovery of divide-and-conquer problem-solving strategies, and c) allows adaptations of human problem-solving strategy with better performance when machine-learned explanations are also presented. 

\end{abstract}

\newpage
\section{Introduction}
\label{sec:intro}
\begin{figure}[H]
	\centering
	\includegraphics[width=0.9\textwidth]{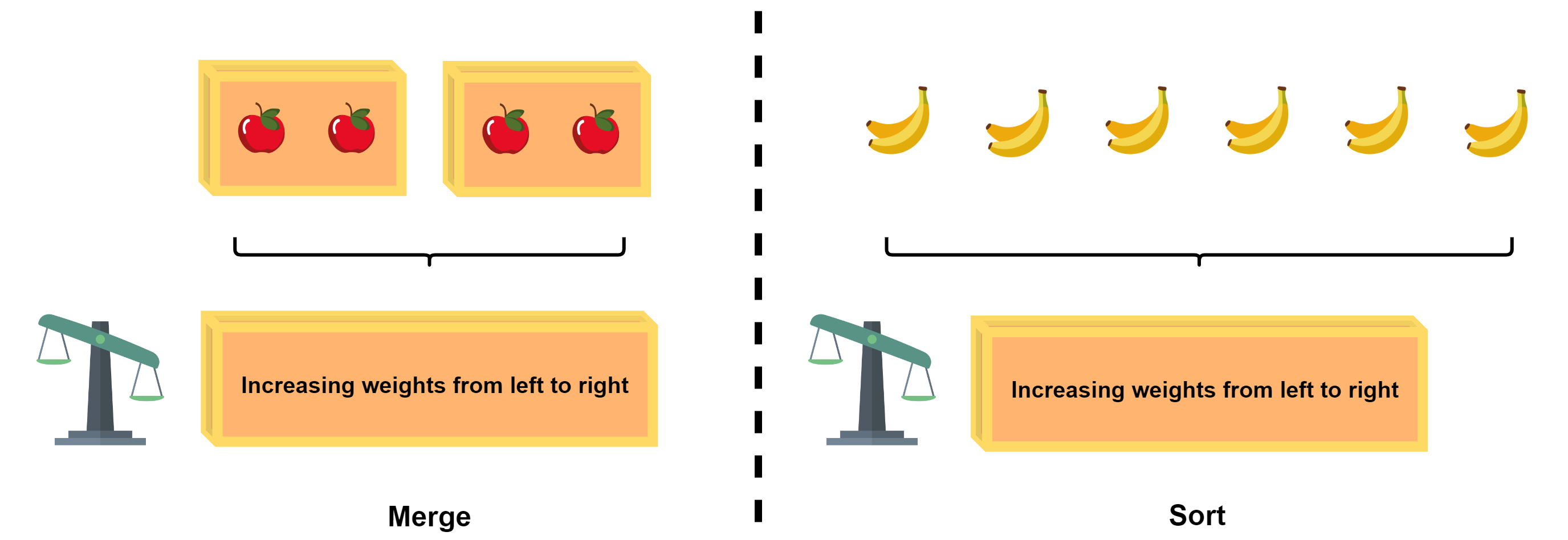}
	\caption{An illustration of teaching materials of merging and sorting that can be learned in incremental order (left to right). All fruits have different weights which can be pairwise compared using the balance scale at the corner. On the left-hand side, two sets of apples should be merged into a larger collection in increasing weights from left to right. On the right-hand side, a set of bananas needs to be arranged in increasing weights from left to right. }
	\label{fig:merge_and_sort_demo}
\end{figure}

Human learning can be described as a concept formation progression in which observations of objects and events are summarised and formulated by some inter-dependent hierarchical structure \cite{concept_formation}. As an example, in the context of teaching algorithms to human students, we may imagine a human teacher using the material in Figure \ref{fig:merge_and_sort_demo} for teaching the merge sort algorithm. In a bottom-up teaching approach with increasing concept complexity, students learn how to merge first (left) and then move on to study sorting (right). In contrast, via a top-down teaching approach which presents concepts with decreasing complexity, students learn to sort first (right) without prior knowledge of merging and study merging afterwards (left). It might be of interest to the human teacher to question which approach yields higher student sorting performance and what effects the two teaching approaches have on students' sorting strategies.
It was stressed by the Machine Learning pioneer Donald Michie \cite{michie_1991_intelligence_human_window} that an intelligent machine not only needs to excel humans in performance but should also possess the capacity to effectively interact with and transfer knowledge to humans. An intelligent machine can in this way fulfil an integrated role that contributes to both the “stock of software” and the “stock of knowledge” \cite{michie_1991_intelligence_human_window}. Owing to increasing awareness of the importance of explainable AI, reported by comprehensive reviews \cite{peek_black_box,systematic_XAI_review,XAI_comprehensive_review}, a great number of studies have emerged with an emphasis on systems that provide support for human understanding. In addition, there is an increasing emphasis on AI that interacts to keep humans ``in the loop'' \cite{miller2019explanation,ute_interactive_learning}. However, owing to the absence of quantifiable definitions and evaluation procedures, some issues such as ambiguity and lack of empirical evaluations have been highlighted \cite{freitas2014,Lipton2018,miller2019explanation}. 

At the intersection of these two paradigms is the research area that focuses on the comprehensibility of machine learning. An operational definition of comprehensibility was provided in \cite{ute2017} for examining machine-human teaching interaction. This definition relates human comprehension to the out-of-sample prediction accuracy of human learners after studying training examples and a machine-learned logic program. It allowed the first demonstration \cite{US2018} of Michie's \textit{Ultra-Strong Machine Learning} (USML) which stressed the role of teaching machine-learned knowledge to humans such that their proficiency on a task can be raised to a higher level with respect to learning from randomly selected examples. Later results in \cite{lun2021} showed that explanations generated from machine-learned logic programs can have both beneficial and harmful effects on human comprehension and suggested, via a cognitive window framework, that explanatory effects are dependent on both descriptive and executional complexity. Despite the classification of explanatory effectiveness in \cite{lun2021}, the effects of the order of concepts in a teaching curriculum on human learning - an essential topic of machine learning and human problem solving that has yet to be explored for understanding the comprehensibility of machine-learned logic programs.

Based on established results and teaching procedures in \cite{lun2021,US2018}, in this work, we extend the frameworks of comprehensibility and explanatory effects to account for the impacts of sequential teaching on comprehension. We refer to the cognitive window framework which attributes the explanatory effects to 1) the descriptive complexity of machine-learned rules and 2) the cost of knowledge application with respect to an execution stack. A human trial of sequential teaching was conducted to examine the effects of concept ordering in curricula and explanations in the context of learning an efficient sorting strategy. We hypothesise the improvement of human comprehension from sequential teaching as the result of the reduction in the size of the hypothesis space associated with the target concept definition. Our proposed framework is consistent with empirical evidence collected from the human trial involving participants with no background in programming. 

\smallskip

\noindent We summarise the paper's contributions as follows:
\begin{itemize}
    \item We define a measure to evaluate beneficial/harmful
	explanatory effects of a sequential teaching curriculum on human comprehension.
	\item We hypothesise the improvement of human comprehension from sequential teaching curricula based on the Blumer bound \cite{blumer_bound}.
	\item We demonstrate based on an analysis of empirical results that a sequential teaching curriculum with increasing concept complexity has a beneficial effect on human comprehension.
	\item We show results that indicate the re-discovery of divide-and-conquer algorithms after human novices learn from concepts with increasing complexity.
	\item We provide evidence of the optimisation of problem-solving strategy as a result of additionally studying explanations generated from machine-learned logic rules.
\end{itemize}

\smallskip

This paper is organised as follows. In Section \ref{sec:related}, we review the literature relevant to the paper. In Section \ref{sec:framework}, we present our theoretical framework of sequential teaching curricula. In Section \ref{sec:exp_framework}, we describe our empirical approach including the experimental hypotheses. In Section \ref{sec:experiments}, we show the results of our main Amazon Mechanical Turk experiment that involved teaching human participants to sort based on the training of merging two ordered sequences. In Section \ref{sec:discussion}, we provide a detailed discussion of our findings on the effects of curriculum order and explanations on human comprehension. In Section \ref{sec:conclusion}, we provide a review of our work and comment on our contributions. We additionally discuss future extensions to the demonstrated results, behavioural cloning and human knowledge re-discovery. 
\section{Related work}
\label{sec:related} 
This section summarises related research on human-machine teaching interactions and highlights our motivations. We give an overview of related results in human learning in Psychology. This is followed by a brief account of learning between humans and machines and objective means for evaluating the comprehensibility of machine explanations.

\subsection{Human learning in curricula with explanations}
The ability of humans to recognise and extract knowledge useful to a problem-solving context is dependent on prior experience and understanding of related domains \cite{simon1979}. While a classification decision is dependent only on information concerning the current state,  goal-directed decision-making requires consideration of a decision's possible effect on future states \cite{barto1989learning}. For sequential decision-making tasks such as the Tower of Hanoi, state transitions of the problem solution can be explicitly expressed by rules \cite{seger1994implicit}. ``Chunking'' reusable sub-goals was demonstrated as an efficient problem-solving approach to encode problem-solution structures and reduce the load on short-term memory \cite{simon1979}. 

The reliance of human problem solving on the problem domain and experience of a person \cite{dienes1999theory} can be attributed to implicit (System 1) and explicit (System 2) knowledge \cite{kahneman2011thinking}. In contrast to implicit knowledge which is only attainable through practice, declarative knowledge can be transferred explicitly in the form of explanations \cite{chi2005complex}. For knowledge accumulation, explicit explanations and hints help make sense of the new information and allow integration into existing knowledge \cite{mayer2004}. In the context of solving simple logical problems \cite{craig1956,kittel1957} and generating LOGO programs \cite{lee1997}, guided learning has been demonstrated more efficient than exploration-oriented approaches. However, multiple studies suggest that explanations should not be presented alone. As demonstrated by \cite{aleven2002effective,anderson1997role,berry1995implicit}, human learners do not always benefit from explanations when the specific problem-solving context is absent. The combination of explicit guidance and worked examples provides a good hands-on problem-solving experience for learners to study and practise for themselves due to the limited capacity of working memory \cite{sweller2007}. The presentation of worked examples along with explanations makes this specific task context available and was shown to support learning \cite{alfieri2011,reed1991use}. 

\subsection{Teaching between the machine and humans}
Machine teaching is an area of machine learning that focuses on the optimisation of training sets and the design of teaching protocols in an example-based setting \cite{machine_teaching_overview}. Studies in machine teaching have provided formal methods for teaching situations in which the goal is to train human students via an optimal set of examples to learn a target hypothesis. The machine teacher in these setups \cite{zhu2019,Rafferty2016,classroom_teaching} is provided information about the quantified cognitive model of the human learner and has access to the learning progress based on the student's responses to examples.  The teacher can then monitor the reactions of students which allows the teacher to make adjustments to teaching materials. However, often excessive assumptions about the teacher’s knowledge are made for deriving theoretical results. The teacher is usually assumed to have perfect knowledge about the computational model of students such as their learning rate and their background knowledge \cite{machine_teaching_limitations}. Thus, owing to the specificity of these theoretical models, they are difficult to apply in real-world teaching settings. 

Communication of machine-learned solutions can be alternatively realised through explanations in the form of rules. For Inductive Logic Programming (ILP) \cite{ILP1991}, systems use logic programs to represent hypotheses as generalisations of examples and background knowledge. ILP systems make use of both induction and abduction, which are critical mechanisms for human knowledge attainment \cite{Hobbs2008,Lemke1967}. A hypothesis learned by an ILP algorithm is represented as a set of rules and can be used to communicate the discovered knowledge to humans. For instance, an ILP system \cite{chess1995} learns optimal strategies of chess endgames for depths 0, 1 and 2 and incorporates a complexity constraint on the number of clauses in a predicate definition based on the hypothesised limitation of working memory capacity of $7\pm2$ ``chunks'' \cite{Miller1956}. A fully automated discovery system \cite{robot_scientist} re-discovers the role of genes of known function by combining ILP-based scientific discovery software with a laboratory robot. In \cite{ILP_chemical_connectivity}, the ILP system produces hypotheses as logic rules that depict the chemical connectivity of molecules and predict mutagenicity, which accommodates collaboration with a human chemist. However, in most systems, it remains an ongoing challenge to ensure human understanding due to the high complexity of information encoded via these interfaces. In a recent work \cite{lun2021}, it was shown that visual and textual explanations for playing a simple two-player game, generated based on rules learned by an ILP system, may cause improvement or degradation in human comprehension after a brief study. The authors demonstrated that the outcome of studying explanations provided is affected by a cognitive window which takes account of the descriptive complexity and execution requirements of the rules.

\subsection{Operational measures of human understanding in AI}
Explainable Artificial Intelligence (XAI) is an area of AI that studies AI systems that allow human understanding by providing human-readable explanations of decisions based on structures and functions. Researchers have attempted to devise AI systems to present knowledge attained in various formats, such as texts \cite{hind2019,textual_explanation_rating}, visualisations \cite{poulin2006,model_agnostic2018,tamagnini2017,neural_image_caption2015} and visualisation-text hybrid demonstrations \cite{schmid2020mutual}. Simpler explanations of local predictions can be devised to provide a localised understanding of a learned classification model \cite{LIME}. For reinforcement learning applications, learned strategies are implicitly encoded by policy functions with continuous domains which are difficult for humans to understand. This concern about the opacity of learned strategies was addressed by incorporating the use of relational biases \cite{robot_interpretable_policy,dvzeroski2001,Zambaldi2019DeepRL} to enhance human understandability. In addition, case-based summaries of a policy \cite{summarisation_2019} from sets of selected states in a larger state space of an agent allow a limited human understanding of the agent's decisions. 

In XAI, there exists a diversity of motivations and technical descriptions of systems which support human understanding. Terms such as transparency, interpretability, explainability and comprehensibility are frequently used terms at the core of XAI \cite{XAI_terms}. Transparency typically refers to the ability of a model or a system to be human-understandable on its own \cite{XAI_terms}. Interpretability relates to the degree of clarity of information revealed to the user by a system \cite{XAI_comprehensive_review}. Explainability stresses the role of a system as an ``explainer'' interface to the user \cite{XAI_terms}. Comprehensibility denotes a system's ability to represent machine-learned knowledge in a human-understandable format \cite{XAI_terms,modeling_comprehensibility}. However, owing to the absence of operational definitions and procedures of evaluation, a great body of recent results was established based on subjective views \cite{freitas2014,miller2019explanation} which are discordant \cite{Lipton2018} with respect to the sub-problem of XAI that the studies attempted to address. Studies were often found to report limited empirical evidence as support of the proclaimed effect and did not take into consideration that human understanding is just as vital as the computational procedures that generate explanatory information \cite{miller2019explanation}. It was stressed in \cite{peek_black_box} that a great body of studies did not evaluate the effect of explanations on their target users. Those works that did include explanatory effect evaluations often failed to provide a good account of the context, results and limitations \cite{systematic_XAI_review}. A number of reports also brought to attention the potential misinterpretation and misuse of explanatory information. For example, partially transparent explanatory information may cause more confusion if key decision-making of systems is omitted \cite{rudin2019}.  The authors in \cite{stumpf2016} expressed trustworthiness concern and showed that human decision-making over-relied on the intelligent system even when explanatory information provided by the system was inaccurate. 

The importance of comprehensibility has long been recognised in machine learning of human-oriented knowledge. Michalski \cite{michalski_inductive_theory} suggested that comprehensible machine learning should produce outputs that share similar structure and semantics as those produced by human experts and should present learned knowledge in human-understandable ``chunks''. Owing to the challenges of quantification, comprehensibility has usually been associated with the complexity of machine-learned models \cite{black_box_survey}. Research on the operational definition of machine learning comprehensibility can be traced back to Michie's definition of \textit{Ultra-Strong} Machine Learning in the 1980s \cite{michie1988}. This criterion stresses machine learners' ability to teach hypotheses to humans such that their performance over the task is improved by a substantial margin compared with human learning alone. An operational definition of comprehensibility \cite{US2018,ute2017} was proposed which assesses human comprehension based on the human out-of-sample classification accuracy. In recent work \cite{lun2021}, both beneficial and harmful effects of explanatory machine learning were identified in the context of symbolic machine learning. However, the effects of the teaching order of concepts on human comprehension have not been fully explored in the literature on machine learning comprehensibility.
\section{Theoretical framework}
\label{sec:framework}
To introduce our theoretical framework, in Section \ref{sec:MIL} we begin by referring to the logic notations and background of Meta-Interpretive Learning (MIL) \cite{MIL_predicate_invention}. A MIL system $Metagol_O$ \cite{MIL_robot_strategy} has been modified in Section \ref{sec:MIL_learning_sorting} to learn a logic theory of sorting as explanations in our empirical human trial. In Section \ref{sec:explanatory_effect_curriculum}, the frameworks of comprehensibility \cite{ute2017} and explanatory effects \cite{lun2021} have been extended to account for sequential teaching curricula containing ``blocks'' of teaching materials. In Section \ref{sec:cognitive_cost}, we refer to the cognitive cost of a logic program \cite{ute2017} which was proposed to approximate the complexity of the human execution of the logic program as humans simulate the knowledge being taught. We describe the cognitive window and sequential teaching improvement conjectures in Section \ref{sec:conjectures} based on \cite{lun2021} and our proposed new framework.

\subsection{Meta-interpretive learning}
\label{sec:MIL}
An Inductive Logic Programming (ILP) \cite{ILP1991} algorithm uses background knowledge $\mathcal{B}$ to induce a hypothesis in the form of a logic program which entails all of the positive examples $\mathcal{E^+}$ and none of the negative examples $\mathcal{E^-}$. Meta-Interpretive Learning (MIL) \cite{MIL_predicate_invention} is a variant of ILP which supports predicate invention \cite{MIL_predicate_invention}, dependent learning \cite{lin2014} and learning of recursive and higher-order programs. A MIL algorithm solves the following problem: given a tuple ($\mathcal{B}$, $\mathcal{M}$, $\mathcal{E^+}$, $\mathcal{E^-}$, $\mathcal{I}$, $\mathcal{H}$) where the background knowledge $\mathcal{B}$ is a first-order logic program, $\mathcal{M}$ is a set of second-order clauses, positive examples $\mathcal{E^+}$ and negative examples $\mathcal{E^-}$ are ground atoms, learns a logic program $H$ such that 
\begin{gather*}
    \forall e^+ \in \mathcal{E^+} \,\,\, \mathcal{M} \, \cup H \, \cup \, \mathcal{B} \models e^+\\
    \forall e^- \in \mathcal{E^-} \,\,\, \mathcal{M} \, \cup H \, \cup \, \mathcal{B} \not\models e^-\\
    \mathcal{M} \, \cup \mathcal{B} \models H
\end{gather*}
$\mathcal{I}$ is a set of predicate symbols reserved for predicate invention. Output logic program $H$ is a hypothesis in the hypothesis space $\mathcal{H}$ described by a finite number of predicate symbols and constants. The background knowledge $\mathcal{B}$ contains a set of predicate definitions as primitives. The set of second-order clauses $\mathcal{M}$ is referred to as meta-rules. Each meta-rule contains existentially quantified second-order variables and universally quantified first-order variables. A MIL algorithm employs meta-rules and substitutes second-order variables with predicate symbols to derive first-order theories as logical generalisations.

\subsection{MIL for learning sorting algorithm}
\label{sec:MIL_learning_sorting}
The MIL system $Metagol$ \cite{cropper2016} is an implementation of a meta-interpreter in Prolog. $Metagol$ supports predicate invention and dependent learning which reduces the size of the hypothesis space and improves learning performance \cite{cropper2017}. $Metagol_O$ \cite{MIL_robot_strategy} makes use of composite objects and actions defined from primitive objects and actions. $Metagol_O$ addresses the issue that most ILP systems only considered the textual complexity of hypothesis programs. The inability to distinguish two algorithms, for instance, insertion sort and quick sort, would result in the less computationally efficient but textually compact algorithm being learned. Both textual and resource costs of hypothesis programs are minimised by $Metagol_O$ using a technique called iterative descent \cite{MIL_robot_strategy} which computes the resource cost of hypotheses of increasing sizes according to a pre-defined cost function. For learning robot postman and robot sorting strategies, the background knowledge is defined in composite actions and objects and provided to $Metagol_O$. The authors showed that $Metagol_O$ converges to hypotheses with the optimal resource complexity and learns resource-efficient robot strategies \cite{MIL_robot_strategy}.

\begin{table}[t]
\centering
	\caption{When the robot performs $parse\_exprs/2$, it parses two expressions by removing the ``<'' symbols, and it puts one sequence of numbers into the left bag and the other sequence of numbers into the right bag. $compare\_nums/2$ first selects a number from the left bag and a number from the right bag. Then it compares the two numbers. Afterwards, $compare\_nums/2$ uses the smaller number to extend the last expression in the memory and puts the larger number back in its original bag. When one of the bags is empty, the robot performs $drop\_bag\_remaining/2$ to append the rest of the numbers to the last expression in the memory. $recycle\_memory/2$ takes all expressions in the memory and fills the expression list $exprs$. $single\_expr/2$ checks if there exists only one expression in the expression list $exprs$. } 
	\begin{tabular}{c|c}
		Definition & Rules\\ 
		& \\\hline
		& \\
		merger/2 & \verb+merger(A,B):-parse_exprs(A,C),merger_1(C,B).+\\
		& \verb+merger_1(A,B):- compare_nums(A,C),merger_1(C,B)+\\
 		& \verb+merger_1(A,B):-compare_nums(A,C),drop_bag_remaining(C,B).+\\ 
 		& \\\hline
 		& \\
		sorter/2 & \verb+sorter(A,B):-merger(A,C),sorter(C,B).+\\
		(after learning  & \verb+sorter(A,B):-recycle_memory(A,C), sorter(C,B).+\\
		merger/2) & \verb+sorter(A,B):-single_expr(A,C), single_expr(C,B).+\\
		& \\\hline
		& \\
		sorter/2 & \verb+sorter(A,B):-parse_exprs(A,C),sorter(C,B).+\\
		(without & \verb+sorter(A,B):-compare_nums(A,C), sorter(C,B).+\\
		learning & \verb+sorter(A,B):-drop_bag_remaining(A,C), sorter(C,B).+\\
		merger/2) & \verb+sorter(A,B):-recycle_memory(A,C), sorter(C,B).+\\
		& \verb+sorter(A,B):-single_expr(A,C), single_expr(C,B).+\\
		& \\
	\end{tabular}
    \label{table:merge_sort_rules}
        \caption{$Metagol_O$ uses two meta-rules. $P$, $Q$, and $R$ are existentially quantified higher-order variables and $x$, $y$, and $z$ are universally quantified first-order variables. $\succ$ is a total ordering constraint over the Herbrand base to guarantee the termination of hypotheses. }
	\begin{tabular}{c|c|c}
		Name & Meta-rule & Orders\\ \hline 
		Chain &  $P(x,y)\leftarrow Q(x,z),R(z,y)$ & $P \succ Q, P \succ R$\\ \hline
		Tailrec & $P(x,y)\leftarrow Q(x,z),P(z,y)$ & $P \succ Q$, \\ 
		\, & \, & $x \succ z \succ y$\\ 
	\end{tabular}
\label{table:meta_rules} 
\vspace{-10pt}
\end{table}
We modify the background knowledge of $Metagol_O$ for learning a bottom-up variant of the merge sort algorithm \cite{bottom_up_merge_sort} for sorting positive integers. Merge sort is a recursive sorting algorithm based on a divide-and-conquer approach. The conventional implementation recursively splits an input sequence into units and performs merging on these units to build up sorted sequences that are shorter. The aforementioned variant takes an input sequence as a list of sub-sequences of length one and iteratively merges two sub-sequences in a bottom-up fashion. For human learners with little or no prior knowledge of recursive sorting algorithms, we consider that iterative algorithms are conceptually easier teaching materials in the case of unfamiliarity with recursive programs. In addition, we assume that human sorting strategies do not usually involve merging. Humans may learn to utilise merging for creative sorting strategies.

A demonstration of the execution of the bottom-up merge sort algorithm is included in Example \ref{ex:bottom_up_variant_merge_sort}. To learn the bottom-up variant of merge sort\footnote{Source code and demos are available on https://github.com/LAi1997/sequential-teaching}, we define the world state that the robot sorter would see. $Metagol_O$ is provided with primitives to manipulate expressions which compactly represent ordered sequences. Let $Lt\_expr$ represent a non-empty monotonically increasing sequence of integers and be defined as $Lt\_expr := \, Int \, | \, Lt\_expr < Lt\_expr$.  

\begin{example}
	Let the sequence \textit{[4, 6, 5, 2, 3, 1]} be an input to the bottom-up variant of merge sort. Merging is iteratively performed from the end of the list with lower indices. Merging take two expressions as inputs, e.g. 4 and 6, and outputs one expression, e.g. \textit{4 < 6}. In \textit{iteration 1}, three merging would be executed on expressions \textit{4} and \textit{6}, \textit{5} and \textit{2}, \textit{3} and \textit{1} which would return \textit{[4 < 6, 2 < 5, 1 < 3]}. In \textit{iteration 2}, one merging would be executed on two leftmost expressions \textit{4 < 6} and \textit{2 < 5} which would give \textit{[2 < 4 < 5 < 6}, \textit{1 < 3}]. In \textit{iteration 3}, expressions \textit{2 < 4 < 5 < 6} and \textit{1 < 3} are merged which would return \textit{[1 < 2 < 3 < 4 < 5 < 6]}. Only one expression remains which represents the sorted sequence and the algorithm terminates with \textit{[1 < 2 < 3 < 4 < 5 < 6]} as output.
	\label{ex:bottom_up_variant_merge_sort}
\end{example}

Objects are separated into primitive objects $O_{prim}$ and composite objects $O_{comp}$ where $O = O_{prim} \cup O_{comp}$ denotes the set of all objects in the robot world. Each composite object is defined by primitive objects or other composite objects. $S$ is a set of states and each state is a tuple of objects. In particular, a state is defined as a tuple $(exprs, energy, left\_bag, right\_bag, memory)$ with an arity of 5. The $exprs$ is a list of $Lt\_expr$. The $energy$ records the resource cost. The $left\_bag$ and $right\_bag$ are lists of the parsed numbers. The $memory$ is a list of newly created $Lt\_expr$. An action $a \in A$ is a function such that $a: S \to S$. Each action is either a primitive action included in $A_{prim}$ or a composite action included in $A_{comp}$. Let A be an enumerable set of actions where $A = A_{prim} \cup A_{comp}$. Every composite action is constructed based on primitive actions or other composite actions. A resource function $r : A \times S \to \mathbb{N}$ defines the resources consumed by carrying out an action $a \in A$ in state $s \in S$. 

$Metagol_O$'s background knowledge is supplied with composite actions $A_{comp}$ = $\{parse\_exprs/2$, $compare\_nums/2$, $single\_expr/2$, $drop\_bag\_remaining/2$, \\ $recycle\_memory/2\}$ for learning $merger/2$ and $sorter/2$ in Table \ref{table:merge_sort_rules}. Meta-rules in Table \ref{table:meta_rules} allow $Metagol_O$ to learn recursive programs. In $merger/2$, two $Lt\_expr$ from $exprs$ are first parsed into $left\_bag$ and $right\_bag$. Then numbers are compared pair-wise from two bags and the smaller number is appended to the last $Lt\_expr$ in $memory$. $sorter/2$ applies $merger/2$ over all $Lt\_expr$ in $exprs$. Constructed $Lt\_expr$ in $memory$ are recycled back to $exprs$ by $recycle\_memory/2$. This process is iterated until there is only one $Lt\_expr$ left in $exprs$ and $memory$ is empty. We define a cost function for $compare\_nums/2$ such that the $energy$ is incremented by one whenever two numbers are successfully compared. We additionally define a constraint over the world states. This constraint based on Spearman's rank correlation coefficient \cite{spearman_rank} ensures the expressions in the current state are not less sorted compared with expressions in the immediate next state.

\subsection{Explanatory effect of curriculum order}
\label{sec:explanatory_effect_curriculum}
We have employed a quantifiable approach that provides operational definitions for the evaluation of the effect of sequential teaching on human comprehension. This allows us to better account for the context, results and limitations. We describe the basis of our framework by referring to previous work on comprehensibility \cite{ute2017} and explanatory effects \cite{lun2021} (Table \ref{table:framework_extension_summary}). Then we define a sequential teaching curriculum and its relative effectiveness with respect to another sequential teaching curriculum. 

Based on an operable definition of comprehensibility \cite{ute2017}, the authors in \cite{lun2021} explored the explanatory effects of machine-learned logic rules on human comprehension and proposed a framework of a cognitive window that focuses on the symbolic subset of machine learning. The definitions of comprehensibility and explanatory effects are extended in this present work to account for sequential teaching curricula in the context of symbolic machine learning. We use the logic programs learned from $Metagol_O$ to generate explanations for our human trial.  For the purpose of experimentation, we only consider curricula with noise-free training examples. This does not mean, however, that the proposed definitions could not be extended outside of the considered scope. 

\begin{table}[t]
    \centering 
    \caption{A summary of the connections between definitions of comprehensibility, explanatory effects and cognitive window in \cite{US2018,lun2021} and the proposed framework to account for effects of sequential teaching curricula. }
    \begin{tabular}{c|c|c}
		 & Definitions and & Extended definitions and   \\ 
		Framework & results in previous works & results for sequential teaching \\ 
		& & \\ \hline
		& & \\
		Comprehensibility & Definitions \ref{def:unaided_comprehension} and \ref{def:machine_explained_comprehension} & Definitions \ref{def:rank_function} to \ref{def:sequential_teaching_comprehension} \\ 
		& & \\ \hline
		& & \\
		Explanatory effects & Definitions \ref{def:explanatory_effect} and \ref{def:beneficial_harmfulness} & Definitions \ref{def:explanatory_effect_sequential_teaching} and \ref{def:beneficial_harmful_sequential_curriculum} \\ 
		& & \\ \hline
		& & \\
		Cognitive window & Definitions \ref{def:term_cost} to \ref{def:solution_cost}, Remark \ref{remark:merge_beneficiality} & Conjecture \ref{conjecture:sequential_teaching_improvement} \\ 
		& and Conjecture \ref{conjecture:cognitive_bound} to \ref{conjecture:cognitive_window} & \\ 
    \end{tabular}
    \label{table:framework_extension_summary}
\end{table}

The operational definition of comprehensibility \cite{ute2017} is an objective means of evaluating human comprehension of a concept based on human out-of-sample predictive accuracy. It was extended in \cite{lun2021} to define the explanatory effectiveness of machine-learned logic programs which measures the extent to which examples with explanations generated from machine-learned logic programs can be simulated by humans. 

\begin{definition} \textbf{(Unaided human comprehension of examples, $C_h$($D$, $H$, $E$))} Given that $D$ is a logic program representing the definition of a target predicate, $H$ is a human group and $E$ is a set of examples of the target predicate. The unaided human comprehension of examples $E$ is the mean accuracy with which a human $h \in H$ after a brief study of $E$ and without further sight can classify new material sampled randomly from the domain of $D$.
\label{def:unaided_comprehension}
\end{definition}

Compared with the unaided human comprehension of examples, the machine-explained human comprehension of examples corresponds to the out-of-sample classification accuracy after studying the machine-learned explanation $M(E)$ where $M$ is a machine learning algorithm and $E$ is a set of training examples.

\begin{definition} \textbf{(Machine-explained human comprehension of examples, $C_{ex}$(\\$D$, $H$, $M(E)$))}: Given that $D$ is a logic program representing the definition of a target predicate, $H$ is a human group, $M(E)$ is a theory learned using machine learning algorithm $M$ and $E$ is a set of examples of the target predicate. The machine-explained human comprehension of examples $E$ is the mean accuracy with which a human $h \in H$ after a brief study of an explanation based on $M(E)$ and without further sight can classify new material sampled randomly from the domain of $D$.
\label{def:machine_explained_comprehension}
\end{definition}

The explanatory effect of a machine-learned theory on human comprehension of a task is defined as the difference between machine-explain and unaided human comprehension of examples of the task. 

\begin{definition} \textbf{(Explanatory effect of a machine-learned theory, $E_{ex}$($D$, $H$, $M(E))$)}: Given a logic program $D$ representing the definition of a target predicate, a human group $H$ and a machine learning algorithm $M$, the explanatory effect of the theory $M(E)$ learned from examples $E$ is
\begin{gather*}
	E_{ex}(D, H, M(E)) = C_{ex}(D, H, M(E)) - C_h(D, H, E)
\end{gather*}
\vspace{-10pt}
\label{def:explanatory_effect}
\end{definition}
In the case that the explanations provided by the machine lead to a positive improvement, the explanatory effect of the learned theory is beneficial to human comprehension. When this difference is negative, the explanatory effect of the machine-learned theory is harmful to human comprehension. 

\begin{definition} \textbf{(Beneficial/harmful effect of a machine-learned theory)}: Given a logic program $D$ representing the definition of a target predicate, a set of training examples $E$, a group of humans $H$, and a symbolic machine learning algorithm $M$:
\begin{itemize}
	\item $M(E)$ learned from $E$ is \textit{beneficial} to $H$ if $E_{ex}(D, H, M(E)) > 0$
	\item $M(E)$ learned from $E$ is \textit{harmful} to $H$ if $E_{ex}(D, H, M(E)) < 0$
	\item Otherwise, $M(E)$ learned from $E$ does not have an effect on $H$
\end{itemize} 
\label{def:beneficial_harmfulness}
\end{definition}

Whether or not an explanatory effect is significantly beneficial or harmful can be determined empirically by applying an independent two-sample t-test or similar method to the means of unaided- and machine-explained human comprehension. In the case that the mean difference is not significant, we cannot classify the effect of a machine-learned theory as either beneficial or harmful. 

The definitions of human comprehension and explanatory effects are extended to account for the ordering of concepts. We define a sequential teaching curriculum to contain a set of concept definitions, sets of examples and explanations. The arrangement of concepts in a curriculum can be decided based on an ordering function which outputs a score to rank concepts. 

\begin{definition} \textbf{(Curriculum rank function, $\lambda$($D$, $E$, $M$))}
Given a set of predicate definitions $Ds$, sets of examples $Es$ and a set of machine learning algorithms $Ms$, $\lambda: Ds \times Es \times Ms \to \mathbb{N}$ is a rank function that returns the rank value of any predicate definition $D \in Ds$ in a curriculum.
\label{def:rank_function}
\end{definition}

A sequential teaching curriculum contains ``chunks'' of teaching material and a curriculum has at least one such ``chunk''. The order of concepts received by humans is determined by a defined rank function which assigns a score to every block of teaching material which contains the concept definitions, examples and explanations.

\begin{definition} \textbf{(Sequential teaching curriculum, $ST$($Ds$, $Es$, $\lambda$, $Ms$))}: Given that $Ds$ = $\{D_1$,$D_2$, ... $D_n\}$ is a set of predicate definitions of size n, $Es$ = $\{E_1$,$E_2$, ..., $E_n\}$ contains sets of examples where $E_i$ are examples of $D_i$ for $1 \leq i \leq n$, $Ms$ denotes a set of machine learning algorithms and $\lambda: Ds \times Es \times Ms \to \mathbb{N}$ is a rank function, a sequential teaching curriculum $ST$($Ds$, $Es$ $\lambda$, $Ms$) is an enumerable set of tuples $\{(R_1, D_1, E_1, M_1), (R_2, D_2, E_2, M_2), ..., (R_n, D_n, E_n, M_n)\}$ in which $R_j = \lambda(D_j, E_j, M_j)$ is the rank of a concept in the curriculum for $1 \leq j \leq n$. A sequential teaching curriculum enumerates concepts with respect to increasing rank value.
\end{definition}

Human comprehension is considered aided by explanations if the machine-learned logic program in a teaching material block is not empty. Otherwise, human comprehension is established based on training examples alone. We proceed to define the effect of a sequential teaching curriculum on the human comprehension of a concept. 

\begin{definition} \textbf{(Human comprehension of examples in a sequential teaching curriculum, $C_{seq}(T, H)$)}: Given that $T = ST(Ds, Es, \lambda, Ms)$ is a sequential teaching curriculum where $Ds$ is a set of n predicate definitions, $Es$ are sets of examples, $\lambda: Ds \times Es \times Ms \to \mathbb{N}$ is a rank function, $Ms$ is a set of machine learning algorithms and $H$ is a human group, the human comprehension of examples in the sequential teaching curriculum is 

\begin{gather*}
C_{seq}(T, H) = \left\{(R_i, \tau) \middle\vert\
\begin{aligned}
& \tau = C_h(D_i, H, E_i), \qquad \qquad \,\, M_i(E_i) = \emptyset\\ 
& \tau = C_{ex}(D_i, H, M_i(E_i)), \qquad M_i(E_i) \not = \emptyset\\
& \qquad \text{for } (R_i, D_i, E_i, M_i) \in S, 1 \leq i \leq n
\end{aligned}
\right\}
\end{gather*}
\label{def:sequential_teaching_comprehension}
\end{definition}

We provide examples of sequential teaching definition which specify a curriculum that presents training materials of merging before introducing training materials of sorting and a curriculum with the reverse order of teaching.   

\begin{example}
    Let $Ds = \{merger, sorter\}$ denote predicate definitions of merging and merge sort, $E_{merge}$ and $E_{sort}$ be the respective example sets and $Es = \{E_{merge}, E_{sort}\}$ and $Ms = \{Metagol_O\}$ which generates explanations. We define $\lambda_{merge/sort}$ ($merger$, $E_{merge}$, $Metagol_O$) = 0 and $\lambda_{merge/sort}$ ($sorter$, $E_{sort}$, $Metagol_O$) = 1. Let $\lambda_{sort/merge}$ ($merger$, $E_{merge}$, $Metagol_O$) = 1 and $\lambda_{sort/mergel}$ ($sorter$, $E_{sort}$, $Metagol_O$) = 0. The merge-then-sort curriculum is represented by $ST$($Ds$, $Es$, $\lambda_{merge/sort}$,$Ms$) which teaches merging before teaching sorting and the sort-then-merge curriculum is denoted by $ST$($Ds$, $Es$, $\lambda_{sort/merge}$,$Ms$) which teaches sorting before teaching merging. $\lambda_{merge/sort}$ specifies the ordering of concepts in the merge-then-sort curriculum $ST$($Ds$, $Es$, $\lambda_{merge/sort}$, $Ms$) and $\lambda_{sort/merge}$ defines the ordering of concepts in the sort-then-merge curriculum $ST$($Ds$, $Es$, $\lambda_{sort/merge}$, $Ms$).
\end{example}

\begin{definition} \textbf{(Effect of a sequential teaching curriculum, $E_{seq}$($C_1$, $C_2$, $D$))}: $C_1$ = $C_{seq}$($ST$($Ds$, $Es$, $\lambda_1$, $Ms$), $H$) and $C_2$ = $C_{seq}$($ST$($Ds$, $Es$, $\lambda_2$, $Ms$), $H$) are sequential teaching curricula where $Ds$ is a set of n predicate definitions and $D \in Ds$ is a predicate definition, $Es = \{E, ...\}$ are sets of examples, $\lambda_1, \lambda_2: Ds \times Es \times Ms \to \mathbb{N}$ are two rank functions, $Ms$ is a set of machine learning algorithms and $H$ is a human group. $R_1 = \lambda_1(D, E, M_1)$ and $R_2 = \lambda_2(D, E, M_2)$ are the ranks of $D$ in $C_1$ and $C_2$ where $M_1, M_2 \in Ms$. The explanatory effect of curriculum $C_1$ over curriculum $C_2$ on predicate definition $D$ is
\begin{gather*}
E_{seq}(C_1, C_2, D) = \tau_1 - \tau_2
\end{gather*}
where $(R_1, \tau_1) \in C_1$, $(R_2, \tau_2) \in C_2$.
\label{def:explanatory_effect_sequential_teaching}
\end{definition}

In Definition \ref{def:sequential_teaching_comprehension}, comprehension measurements are obtained with respect to the ranking of concepts in a sequential teaching curriculum. The effect of a sequential teaching curriculum reflects the degree to which human comprehension of a concept is affected by a curriculum. Given two sequential teaching curricula and a concept of interest, the effect of one curriculum over another curriculum corresponds to the difference in comprehension measurements of the concept. A significant difference in human comprehension as a result of learning from one curriculum over another curriculum can be classified as a beneficial or harmful effect of learning from the former curriculum.

\begin{definition} \textbf{(Beneficial/harmful effect of a sequential teaching curriculum)}: Given that $C_1$ = $C_{seq}$($ST$($Ds$, $Es$, $\lambda_1$, $Ms$), $H$) and $C_2$ = $C_{seq}$($ST$($Ds$, $Es$, $\lambda_2$, $Ms$), $H$) are sequential teaching curricula where $Ds$ is a set of n predicate definitions and $D \in Ds$ is a predicate definition, $Es$ are sets of examples, $\lambda_1, \lambda_2: Ds \times Es \times Ms \to \mathbb{N}$ are two rank functions, $Ms$ is a set of machine learning algorithms, and $H$ is a human group, the curriculum $C_1$ in comparison with the curriculum $C_2$ is
\begin{itemize}
	\item \textit{beneficial} to $H$ on $D$ if $E_{seq}(C_1, C_2, D) > 0$
	\item \textit{harmful} to $H$ on $D$ if $E_{seq}(C_1, C_2, D) < 0$
	\item Otherwise there is no effect on $H$ for $D$ in curriculum $C_1$ compared with curriculum $C_2$
\end{itemize} 
\label{def:beneficial_harmful_sequential_curriculum}
\end{definition}

The effect on the predicate $D$ is beneficial when the curriculum ordered by $\lambda_1$ results in a better comprehension of $D$ than the comprehension of $D$ in the curriculum ordered by $\lambda_2$. Conversely, the effect on $D$ is harmful in the case that significant degradation of comprehension is observed when comparing the comprehension of $D$ between curricula described by $\lambda_1$ and $\lambda_2$. As with the effect of a machine-learned theory (Definition \ref{def:beneficial_harmfulness}), whether or not an effect of a sequential teaching curriculum is significantly beneficial or harmful can be determined using an independent two-sample t-test or similar method.

\subsection{Cognitive cost of a logic program}
\label{sec:cognitive_cost}
A consensus between the literature on cognitive psychology and artificial intelligence \cite{Miller1956,Johnson_laird1986,Newell1990} is that the information processing of humans in working memory can be modelled by manipulation of symbols in machines which can be formally captured by Turing machines. The limitation on working memory capacity corresponds to a bounded tape length and instruction complexity in Turing Machines. For rule-based concept acquisition \cite{Bruner1956}, human concept attainment is analogical to a search in a collection of hypotheses guided by some preference ordering which is comparable to version space learning in machine learning \cite{Mitchell1982}. In addition, the organisation of complex action sequences into hierarchical structures in human information processing is an efficient mechanism for general problem-solving. In the contexts of analogical problem-solving \cite{Carbonell1985} and production systems \cite{Newell1990}, top-level goals are decomposed into sub-goals which can be described by a set of rules that guide actions in problem-solving sequences. Although rules generally can be considered as procedural knowledge, in complex domains human verbalisation of rules utilises declarative memory \cite{Anderson1993,Schmid11}. The explicit representations of a human problem-solving solution correspond to rule-like descriptions. For the following definitions, we assume that:
\begin{itemize}
    \item Human learners are version space learners who can make only limited hypothesis space searches
    \item Human learning is guided by meta-rules which construct the sub-goal structure and uses predicates as background knowledge
    \item An increase in the execution complexity of the problem solution can have a negative effect on performance
    \item Rules can be declaratively represented in a verbalisable form whose complexity can be measured by the Kolmogorov complexity
\end{itemize}  

The cognitive cost is a variant of Kolmogorov complexity \cite{Kolmogorov1963} proposed in \cite{lun2021}. The cognitive cost estimates the textual complexity of logic terms and atoms declared in the execution stack of working memory. These logic terms and atoms are instances of predicates in a datalog program during the execution of a goal. 

A datalog program is a declarative subset of logic programs that represent data structures using predicate arguments but do not use function symbols. Given logic terms and atoms $T_1,T_2, ..., T_{n-1}, T_n$, a tuple of finite arity can represented as
\begin{gather*}
tuple(T_1, tuple(T_2, tuple(..., tuple(T_{n-1}, tuple(T_n, \epsilon))...)))
\end{gather*}
Similarly, a list of finite size can be represented as
\begin{gather*}
    list(T_1, list(T_2, list(..., list(T_{n-1}, list(T_n, \epsilon))...)))
\end{gather*}
where $\epsilon$ denotes an empty character. The cost of logic terms and atoms is computed based on the length of the representation, in particular, the number of symbols involved.  

\begin{definition} \textbf{(Cognitive cost of a logic term and atom, $C(T)$)}: Given $T$ a logic term or atom, the cost of $C(T)$ can be computed as follows:
\begin{itemize}
	\item $C(\top) = C(\bot) = 1$
	\item A variable $V$ has cost $C(V) = 1$ 
	\item A constant $c$ of a finite length has cost $C(c)$ which is the number of digits and characters in $c$
	\item An atom $Q(T_1,T_2,...)$ of a finite arity has cost $C(Q(T_1,T_2,...)) = 1 + C(T_1) + C(T_2) +$ … 
\end{itemize}
\label{def:term_cost}
\end{definition}

\begin{example}
	An atom $list(1,list(0, \epsilon))$ which represents a list [1, 0] has a cognitive cost $C(list(1,list(0, \epsilon))) = 5$.
\end{example}

\begin{example}
	An atom $merger(State1, State2)$ with variables $State1$ and $State2$ has a cognitive cost $C(merger(State1, State2)) = 3$.
\end{example}

In the event of a query $q$, an execution stack resembles the mental computation and represents the memory of grounded logic terms and atoms instantiated.

\begin{definition} \textbf{(Execution stack of a datalog program, $S(P, l, q)$)}: Given a query $q$ and a non-negative integer $l$, the execution stack $S(P, l, q)$ of a datalog program $P$ is a finite set of size $l$ of atoms or terms evaluated during the execution of $P$ to compute an answer for $q$. An evaluation in which an answer to the query is found ends with value $\top$, and an evaluation in which no answer to the query is found ends with $\bot$. 
\label{def:stack}
\end{definition}

The cognitive cost represents the tax of memorising goals in the human short-term memory. The effort of maintaining goals in working memory correlates to the performance of human problem-solving \cite{Carpenter1990}. The complexity of executing a set of rules is estimated by the sum of the cognitive cost of logic terms and atoms in the execution stack. The depth bound of an execution stack denotes the limitation of working memory and ensures the computation of the sum of cognitive costs halts.

\begin{definition} \textbf{(Cognitive cost of a datalog program, $Cog(P, q)$)}: Given a query $q$, a non-negative integer $l$, the cognitive cost of a datalog program $P$ is
	\begin{gather*}
	Cog(P, q) = \sum_{t \in S(P, l, q)} C(t)
	\end{gather*}\label{def:program_cost}
	\vspace{-10pt}
\end{definition}

\begin{example}
	The program $merger/2$ in Table \ref{table:merge_sort_rules} takes two input $Lt\_expr$ expressions in the expression list of a world state and produces an expression $Lt\_expr$ in the memory of the output world state. The output expression contains integers of increasing magnitude from left to right connected by the ``<'' symbol. Given $ex_1 = list(1,list(0, \epsilon))$, $lb_1 = \epsilon$, $rb_1 = \epsilon$ and $m_1 = \epsilon$, $s_1$ = $tuple$($ex_1$, $tuple$(0, $tuple$($lb_1$,$tuple$($rb_1$,$m_1$)))) denotes the initial world state and $C(s_1) = 13$. Let $ex_2$ = $list$($\epsilon$, $\epsilon$), $lb_2 = list(1, \epsilon)$, $rb_2$ = $list$(0, $\epsilon$) and $m_2 = \epsilon$. $s_2$ = $tuple$($ex_2$, $tuple$(0, $tuple$($lb_2$,$tuple$($rb_2$,$m_2$)))) denotes the world state after executing $parse\_expr/2$ and $C(s_2) = 15$. $parse\_expr/2$ takes two expressions from the expression list and puts one expression into the left bag and puts the other into the right bag. Let $V_1$ and $V_2$ be variables. The length of the execution stack denoted by $S(merger, 4, merger(s1, V_1))$ is 4. The cognitive cost of executing $merger/2$ given the query $merger(s_1, V_1)$ is 60.
	\vspace{-15pt}
	\begin{table}[H]
		\centering
		\begin{tabular}{ c | c }
			$S(merger, 4, merger(s1, V_1))$ & $C(T)$ \\ \hline 
			$merger(s_1, V_1)$ & 15 \\ \hline
			$parse\_expr(s_1, V_2)$ & 15 \\ \hline
			$parse\_expr(s_1, s_2)$ & 29 \\ \hline
			$\bot$ & 1 \\ \hline
			$Cog(merger, merger(s_1, V_1))$ & \textbf{60}
		\end{tabular}
		\vspace{-10pt}
	\end{table}
\end{example}

In human problem solving, the role of background knowledge is to facilitate the transfer of existing solutions \cite{Anderson1989,Novick1991} to the current context. In the case that auxiliary knowledge is absent, the construction of the solution relies on lower-level sub-goals and performance is limited by human operational error. A primitive problem solution thus denotes a program that involves the minimum amount of background knowledge of the task. 

\begin{definition} \textbf{(Minimum primitive solution program, $\bar{M}(\phi,E)$)}: Given a set of primitives $\phi$ and examples $E$, a datalog program learned from examples $E$ using a symbolic machine learning algorithm $\bar{M}$ and a set of primitives $\phi' \subseteq\phi$ is a minimum primitive solution program $\bar{M}(\phi,E)$ if and only if for all sets of primitives $\phi'' \subseteq\phi$ where $|\phi''| < |\phi'|$ and for all symbolic machine learning algorithm $M'$ using $\phi''$, there exists no machine-learned program $M'(E)$ that is consistent with examples $E$. 
\label{def:min_solution}
\end{definition}

\begin{definition} \textbf{(Cognitive cost of a problem solution, $CogP(E, \bar{M}, \phi, q)$)}: Given examples $E$, primitive set $\phi$, a query $q$ and a symbolic machine learning algorithm $\bar{M}$ that learns a minimum primitive solution, the cognitive cost of a problem solution is
\begin{gather*}
CogP(E, \bar{M}, \phi, q) = Cog(\bar{M}(\phi, E), q)
\end{gather*}
where $\bar{M}(\phi, E)$ is a minimum primitive solution program.
\label{def:solution_cost}
\end{definition}

The presence of informative knowledge enables efficient problem-solving as projections from background knowledge and allows shortcuts in the construction of the current solution process. In comparison with a primitive problem solution, executional shortcuts in a non-primitive solution could lead to less cognitive load on working memory and fewer performance errors during the execution of the solution. 

\subsection{Conjectures of human comprehension}
\label{sec:conjectures}
Donald Michie \cite{michie_human_window} discussed the effect of the representational and executional cost of a program on human understanding. He described the notion of a human window in terms of a class of programs with the right balance of storage and computational complexity that are fit for human understanding. The cognitive window framework \cite{lun2021} shares Michie's view of the human window and describes two constraints of machine-learned theory in regard to its effects on human comprehension. The first constraint relates to the size of the hypothesis space associated with the representation of a machine-learned logic theory. This constraint specifies the limitation of human search in the hypothesis space.

\begin{conjecture} 
	\textbf{(Cognitive bound on the hypothesis space size, $B(P, H)$)}: Consider a symbolic machine-learned datalog program $P$ using $p$ predicate symbols and $m$ meta-rules each having at most $j$ body literals. Given a group of humans $H$, $B(P,H)$ is a population-dependent bound on the size of hypothesis space such that at most $n$ clauses in $P$ can be comprehended by all humans in $H$ and $B(P,H)=m^n p^{(1+j)n}$.
	\label{conjecture:cognitive_bound}
\end{conjecture}

The conjecture of the human searchable hypothesis space size $B(P, H)$ relates to the MIL complexity analysis in \cite{cropper2017,lin2014}. The authors in \cite{lun2021} demonstrated empirical evidence which supports Conjecture \ref{conjecture:cognitive_bound}. Humans learned can only learn some of the rules for a two-players simple game when the descriptive size of the rules is high. This implies that the hypothesis space associated with the program class is too large for a complete search and imposes a high cognitive load on working memory. Since only a fraction of the original rules is learned, problem solutions which produce action sequences are not complete and errors become more likely to occur. 

The cognitive window conjecture describes that a machine-learned theory has a harmful explanatory effect on comprehension when learning requires a search in a hypothesis space that is too large for working memory to manage. In addition, a machine-learned theory has no beneficial explanatory effect if its cognitive complexity in executing explained knowledge is not lower than the cognitive complexity of a solution that uses less auxiliary knowledge.

\begin{conjecture}
    \textbf{(Cognitive window of a machine-learned theory)}: Given a logic program $D$ representing the definition of a target predicate, a machine learning algorithm $M$, a minimum primitive solution learning algorithm $\bar{M}$ and examples $E$, $M(E)$ is a machine-learned theory using the primitive set $\phi$ and belongs to the program class with hypothesis space $S$. For a human group $H$,  $E_{ex}$ satisfies both
    \begin{enumerate}
        \item $E_{ex}(D, H, M(E))$ < $0$ if $|S|$ > $B(M(E), H)$ 
        \item $E_{ex}(D, H, M(E)) \, \leq \, 0$ if $Cog(M(E), x) \, \geq \, CogP(E, \bar{M}, \phi, x)$ for all unseen query $x$ that $h \in H$ have to perform after study
    \end{enumerate}
\label{conjecture:cognitive_window}
\end{conjecture}

Explanations are only beneficial if they are of appropriate complexity and are not overwhelming nor more cognitively expensive than a primitive solution to the problem. This cognitive window suggests that complex machine-learned models and models which cannot provide abstract descriptions are difficult to be comprehended by humans effectively.

\begin{remark}
$Metagol_O$ is provided with a set of composite predicate definitions $A_{comp} = \{parse\_exprs/2, compare\_nums/2, single\_expr/2, drop\_bag\_remaining/2, \\ recycle\_memory/2\}$ to learn $merger/2$ in Table \ref{table:merge_sort_rules}. $A_{comp}$ is constructed from a set of primitive action definitions $A_{prim}$. The composite actions of breaking number sequences into individual numbers ($parse\_exprs/2$), comparing two numbers ($compare\_nums/2$) and appending two sorted sequences ($drop\_bag\_remaining/2$) are sufficient for merging and the respective predicates appear in the learned logic program $merger/2$. Given action definitions $A = A_{prim} \cup A_{comp}$, since composite actions $A_{comp}$ are constructed from multiple primitive actions from $A_{prim}$, $A_{comp}$ is the necessary and sufficient subset of $A$ to learn a definition of merging. This implies that $merger/2$ learned by $Metagol_O(E)$ is the minimum primitive solution. Therefore, given test examples $X$ of merging and sufficient training examples $E$, $Cog$($merger$, $x$) = $CogP$($E$, $Metagol_O$, $A$, $x$) for all $x \in X$.
\label{remark:merge_beneficiality}
\end{remark}

The complexity analysis based on the Blumer bound \cite{blumer_bound} in \cite{playgol} attributes a sample complexity decrease to the introduction of new predicate symbols to the machine learner. Previously learned predicate definitions extend the background. Reusing learned concepts leads to a decrease in the size of the target hypothesis and a reduction in the size of the hypothesis space which improves learning performance \cite{playgol}. Based on our assumptions in Section \ref{sec:cognitive_cost}, we consider an analogy between the human reusing of previously learned concepts and the introduction of predicate definitions into a new learning context. By referring to Definition \ref{def:beneficial_harmful_sequential_curriculum}, we attribute a beneficial effect of a curriculum on human comprehension over another curriculum to an improvement in the sample complexity based on the analysis in \cite{playgol}.

\begin{conjecture}
    \textbf{(Sequential teaching curriculum improvement)}: $C_1$ = $C_{seq}$($ST$(\\$Ds$, $Es$, $\lambda_1$, $Ms$), $H$) and $C_2$ = $C_{seq}$($ST$($Ds$, $Es$, $\lambda_2$, $Ms$), $H$) are sequential teaching curricula where $Ds$ is a set of n predicate definitions and $D \in Ds$ is a predicate definition, $Es$ are sets of examples, $\lambda_1, \lambda_2: Ds \times Es \times Ms \to \mathbb{N}$ are rank functions, $Ms$ is a set of machine learning algorithms and $H$ is a human group. $R_1$, $R_2$ denote ranks of $D$ such that $(R_1, D) \in C_1$ and $(R_2, D) \in C_2$. For $1 \leq i \leq n$ such that $(R_i, D_i) \in C_1$ and $R_i < R_1$, let $p$ denote the sum of the number of predicates in $D_i$ and u be the minimum number of clauses to express $D$ using $D_i$. For $1 \leq j \leq n$ such that $(R_j, D_j) \in C_2$ and $R_j < R_2$, let $p + c$ denote the number of predicates in $D_j$ and $u + k$ be the minimum number of clauses to express $D$ using $D_j$. $E_{seq}(C_1, C_2, D) > 0$ when:
    \begin{gather*}
    u \ln(p) < (u+k)\ln(p+c)
    \end{gather*}
    \label{conjecture:sequential_teaching_improvement}
    \vspace{-10pt}
\end{conjecture}
Given a target predicate $D$ and a fixed human search bound $B(D, H)$ for a human group $H$, human learners are more likely to find $D$ if the hypothesis space associated becomes smaller as a result of learning new knowledge prior to learning $D$. For a fixed set of training examples in two curricula, a reduction in sample complexity correlates to a decrease in predictive error and an improvement in performance.
\section{Experimental framework}
\label{sec:exp_framework}
In previous works \cite{lun2021,US2018}, human comprehension was measured by the accuracy with which participants answered classification questions. The accuracy of problem-solving performance for classification tasks and sequential decision-making tasks can be different due to the nature of the operations involved. In this section, we present our approach to evaluating human sorting performance based on statistics and describe our experimental hypotheses in relation to the theoretical framework.

\subsection{Evaluating human comprehension of sorting}
For merging and sorting, human comprehension corresponds to the competence of participants to arrange items with respect to the total ordering of the items. Decisions related to the arrangement of items are reflected by the classification of items into respective positions in the sequence. While the out-of-sample classification accuracy is an objective measure of human comprehension for classification tasks, it does not estimate the error of sequential predictions. Although alternative metrics exist, we use Spearman's rank correlation coefficient \cite{spearman_rank} as a means to assess the error between the expected sequence and the actual sequence provided by a participant. Spearman's rank correlation coefficient is a non-parametric rank correlation test and indicates the degree of monotonic ordering between two variables. Given two sequences of values representing samples of two variables, when Spearman's rank correlation coefficient is $+1$ or $-1$, it means the variables are perfectly monotonic functions of each other. On the other hand, when Spearman's rank correlation coefficient tends to $0$, it indicates there is no monotonic relationship between the two variables. We use Spearman's rank correlation coefficient to examine the extent of monotonic alignment between a sequence of integers that has been provided by a human and the correctly arranged sequence produced by a machine algorithm.

In addition, we take into consideration the possibility that participants provide incorrect answers that are inversely sorted with respect to the order specified in the experiment instructions. In this case, a coefficient score tending to $-1$ would mean that the participant successfully managed to sort the sequence but forgot the specified ordering. In the experiment trials, we found that such answers rarely occurred. However, in order to avoid diluting the significance of results due to the inclusion of negative coefficients, we define the following function to discount negative coefficients and normalise all of them into the interval [0, 1]. 

\begin{definition} (\textbf{Normalised performance score, $perf(R_h,R_m)$})
Given two sets of integers $R_h$ and $R_m$, a defined discounting constant $0 < c < 1$ and the Spearman's rank correlation coefficient $\rho$, the normalised performance score is defined as 
\begin{gather*}
    \begin{aligned}
        	perf(R_h,R_m) = 
        	\begin{cases}
        	& \rho(R_h,R_m), \qquad \qquad \rho(R_h,R_m) \geq 0 \\
	        & |\rho(R_h,R_m)| \cdot c, \qquad \, \rho(R_h,R_m) < 0
	       \end{cases}
    \end{aligned}
\end{gather*}
\label{def:performance_score}
\end{definition}

Example \ref{ex:performance_score} shows the performance evaluation of two unordered sequences by comparing them to a sorted sequence.

\begin{example}
Let $R_{h1}$ denote the sequence \textit{[4, 6, 5, 2, 3, 1]}, $R_{h2}$ denote the sequence \textit{[1, 2, 6, 3, 4, 5]} and $R_m$ be the perfectly monotonic sequence \textit{[1, 2, 3, 4, 5, 6]} sorted by a machine program. Compared with $R_m$, $R_{h2}$ has only one number that is out of place. However, $R_{h1}$ is almost a sequence with decreasing magnitude with numbers 2 and 4 misplaced. Given $R_m$ is a monotonically increasing sequence, $R_{h2}$ aligns better with the machine output $R_m$ and $perf(R_{h1}, R_m) = .386 \, < \, perf(R_{h2}, R_m) = .657$.
\label{ex:performance_score}
\end{example}

The Spearman's rank correlation coefficient lies in the interval [-1, 1] and the absolute value of the negative coefficient is multiplied by a discounting constant. Based on results from multiple trials, the discounting constant $c$ is set to .5 to moderately penalise responses that involve reversed sorted sequences.

\subsection{Experimental hypotheses}
\begin{table}[t]
\centering
\caption{We consider two independent experimental variables. The order of concepts in the curriculum (\textbf{CO}) is varied between learning merging before learning sorting (\textbf{MS}) and learning sorting before learning merging (\textbf{SM}). The presence of explanations generated from machine-learned rules (\textbf{EX}) alters between learning with explanations (\textbf{WEX}) and learning without explanations (\textbf{WOEX}). The performance score of human participants (\textbf{PS}) is the dependent experimental variable.}
	\begin{tabular}{c|c|c}
		Variable name & Variable type & Treatment abbreviations \\ & & \\
		\hline
		& & \\
		Curriculum order (\textbf{CO}) & Independent variable & \textbf{MS}, \textbf{SM}\\ 
		& & \\ \hline
		& & \\
		Presence of explanations (\textbf{EX}) & Independent variable & \textbf{WEX}, \textbf{WOEX} \\ & & \\\hline
		& & \\
		Performance score (\textbf{PS}) & Dependent variable & --\\
		& & \\
	\end{tabular}
\label{table:experimental_variables} 
\end{table}

We define our experimental hypotheses in this section. Table \ref{table:experimental_variables} shows the experimental variables and treatments. We examine the effect of the curriculum order (\textbf{CO}) via two teaching arrangements: a) learning merging before learning sorting (\textbf{MS}) and b) learning sorting before learning merging (\textbf{SM}). In addition, we assess the effect of the presence of explanations (\textbf{EX}) in two environments: a) learning with explanations of merging (\textbf{WEX}) b) learning without explanations of merging (\textbf{WOEX}). 

Explanations are not provided for learning sorting to avoid introducing merge prematurely in the curricula that learn sorting before learning merging. We use the normalised performance score $perf$ in Definition \ref{def:performance_score} to measure both human performances of merging and sorting (\textbf{PS}). 

\begin{table}[t]
    \centering 
    \caption{Definitions of the rank functions. In curricula \textbf{MS/WEX} and \textbf{MS/WOEX}, humans learn merging before learning sorting. In curricula \textbf{SM/WEX} and \textbf{SM/WOEX}, humans learn sorting before learning merging. The rank functions of \textbf{MS} assign a lower rank value to merging. The rank functions of \textbf{SM} assign a lower rank value to sorting. In curricula \textbf{MS/WEX} and \textbf{SM/WEX}, explanations about merging are generated from $Metagol_O$. }
    \begin{tabular}{c|c}
		Curricula abbreviation & Rank function definitions \\
		& \\\hline
		& \\
		\textbf{MS/WEX} & $\lambda_{MS/WEX}$($merger$, $E_{merge}$, $Metagol_O$) = 0\\
		& $\lambda_{MS/WEX}$($sorter$, $E_{sort}$, $\emptyset)$ = 1\\
		& \\ \hline
		& \\
		\textbf{MS/WOEX} & $\lambda_{MS/WOEX}$($merger$, $E_{merge}$, $\emptyset$) = 0\\
		& $\lambda_{MS/WOEX}$($sorter$, $E_{sort}$, $\emptyset)$ = 1 \\
		& \\ \hline
		& \\
		\textbf{SM/WEX} & $\lambda_{SM/WEX}$ ($merger$, $E_{merge}$, $Metagol_O)$ = 1\\
		& $\lambda_{SM/WEX}$ ($sorter$, $E_{sort}$, $\emptyset)$ = 0 \\
		& \\ \hline
		& \\
		\textbf{SM/WOEX} & $\lambda_{SM/WOEX}$ ($merger$, $E_{merge}$, $\emptyset)$ = 1\\
		& $\lambda_{SM/WOEX}$ ($sorter$, $E_{sort}$, $\emptyset) = 0$ \\
		& \\ 
    \end{tabular}
    \vspace{-10pt}
    \label{table:rank_functions}
\end{table}

Let $Ds = \{merger, sorter\}$ denote target theories of merging and sorting. Let $Es = \{E_{merge}, E_{sort}\}$ where $E_{merge}$ denotes a sufficient set of examples for learning the target theory of merging and $E_{sort}$ is a sufficient set of examples for learning the target theory of sorting. Both human learners and $Metagol_O$ are provided with the same sets of examples $Es$. Let $Ms = \{Metagol_O, \emptyset\}$ denote the machine learning algorithms used to learn rules for generating explanations and $\emptyset$ is an empty program. $H$ stands for a human group. The rank functions for curricula that learn merging before learning sorting are defined as $\lambda_{MS/WEX}$ and $\lambda_{MS/WOEX}$. The rank functions of curricula that learn merging before learning sorting are defined as $\lambda_{SM/WEX}$ and $\lambda_{SM/WOEX}$. The definitions of the rank functions are summarised in Table \ref{table:rank_functions}.

Let $C_{MS/WEX}$ = $C_{seq}$($ST$($Ds$, $Es$, $\lambda_{MS/WEX}$, $Ms$), $H$) denote the human comprehension of the sequential teaching curriculum that learns merging before learning sorting with explanations of merging. Let $C_{MS/WOEX}$ = $C_{seq}$($ST$($Ds$, $Es$, $\lambda_{MS/WOEX}$, $Ms$), $H$) denote the human comprehension of the sequential teaching curriculum that learns merging before learning sorting without explanations of merging. Let $C_{SM/WEX}$ = $C_{seq}(ST(Ds, Es,$ $\lambda_{SM/WEX}$, $Ms$), $H)$ denote the human comprehension of the sequential teaching curriculum that learns sorting before learning merging with explanations of merging. Let $C_{SM/WOEX}$ = $C_{seq}(ST(Ds, Es,$ $\lambda_{SM/WOEX}$, $Ms$), $H)$ denote the human comprehension of the sequential teaching curriculum that learns sorting before learning merging without explanations of merging. We then introduce hypotheses that focus on how curriculum order and explanations generated from machine-learned rules affect human learning of sorting. In Table \ref{table:experimental_hypothesis_summary}, we show how the following experimental hypotheses relate to the conjectures and our framework. 

\begin{table}[t]
    \centering 
    \caption{A summary of experimental independent variables, definitions and results related to each experimental hypothesis. }
    \begin{tabular}{c|c|c}
		Hypotheses & Independent experimental variables & Related definitions \\ 
		& & \\\hline
		& & \\
		H1 & \textbf{CO} & \\ 
		H2 & \textbf{EX} & Definition \ref{def:rank_function} to \ref{def:beneficial_harmful_sequential_curriculum}, \ref{def:performance_score} and Conjecture \ref{conjecture:sequential_teaching_improvement}\\ 
		H3 & \textbf{CO} and \textbf{EX} &  \\ 
		& & \\ \hline
		& & \\
		H4 & \textbf{EX} & Definition \ref{def:unaided_comprehension} to \ref{def:beneficial_harmfulness}, \ref{def:performance_score} and Conjecture \ref{conjecture:cognitive_window} \\ 
		& & \\ \hline
		& & \\
		H5 & \textbf{EX} & Definition \ref{def:performance_score} \\ 
		& & \\
    \end{tabular}
\label{table:experimental_hypothesis_summary}
\end{table}

\bigskip

\noindent \textbf{H1}: \textit{Learning merging before learning sorting leads to a beneficial effect on human comprehension of sorting ($E_{seq}$($C_{MS/WEX}$, $C_{SM/WEX}$, $sorter$) + $E_{seq}$(\\$C_{MS/WOEX}$, $C_{SM/WOEX}$, $sorter$) > 0) with respect to learning sorting before learning merging}. \hangindent=20pt 

\bigskip

In \textbf{H1}, we examine whether human learners achieve better sorting performance from the curriculum with increasing concept complexity (\textbf{MS}) than from the curriculum with decreasing concept complexity (\textbf{SM}).

\bigskip

\noindent \textbf{H2}: \textit{Learning merging with explanations results in a beneficial explanatory effect on human comprehension of sorting ($E_{seq}$($C_{MS/WEX}$, $C_{MS/WOEX}$,$sorter$) + \\ $E_{seq}$($C_{SM/WEX}$, $C_{SM/WOEX}$,$sorter$) > 0) with respect to learning merging without explanations}. \hangindent=20pt 

\bigskip

In \textbf{H2}, we assess if learning merging with aids in the form of explanations constructed from machine-learned rules (\textbf{WEX}) improves sorting performance compared with learning merging without explanations (\textbf{WOEX}).

\bigskip

\noindent \textbf{H3}: \textit{Learning merging with explanations further increases the beneficial effect of curriculum order on human comprehension of sorting ($E_{seq}$($C_{MS/WEX}$, $C_{MS/WOEX}$, $sorter$) - $E_{seq}$($C_{SM/WEX}$, $C_{SM/WOEX}$, $sorter$) > 0) with respect to learning merging without explanations}. 
\hangindent=20pt 

\bigskip

In \textbf{H3}, we inspect if there is an interaction effect between curriculum order (\textbf{CO}) and explanations (\textbf{EX}) on sorting performance. 

\bigskip

We suspect that for human learners who have no previous programming experience, learning merging in the presence of explanations results in adapting new sorting strategies in comparison with learning merging in the absence of explanations. Since merging can be taught in isolation and it does not depend on the knowledge of sorting in the curricula, we refer to Definition \ref{def:beneficial_harmfulness} and the cognitive window to account for the effect of explanations on merging. However, owing to $Metagol_O$ learning a minimum primitive solution $merger/2$, we anticipate no beneficial effect on human comprehension of merging as a result of learning from explanations.  Let $A$ denote the background knowledge of $Metagol_O$ containing definitions of primitive and composite actions and $X$ denote questions that the human group $H$ performs in the merge performance test. 

\bigskip

\noindent \textbf{H4}: \textit{Learning merging with explanations generated from rules without a low cognitive cost $(Cog(merger, x)$ $\geq$ $CogP$($E_{merge}$, $Metagol_O$, $A$, $x$) for all $x \in X$) does not result in a beneficial explanatory effect on human comprehension of merging ($E_{ex}$($merger$, $H$, $merger$) $\leq 0$)}. 
\hangindent=20pt 

\bigskip

In \textbf{H4}, we examine if learning with explanations of merging (\textbf{WEX}) results in no observable performance difference or worse performance with respect to learning without explanations of merging (\textbf{WOEX}).

\bigskip

\noindent \textbf{H5}: \textit{Learning merging with explanations before learning sorting leads to adaptation of efficient human sorting strategies with better performance}.
\hangindent=20pt 

\bigskip

We consider executed comparisons of human participants as problem-solving traces. In Section \ref{sec:results}, we estimate through human trace analysis the correspondence between human sorting strategies and machine sorting algorithms in training and performance tests. An underlying assumption is that learning a new strategy during the performance test is challenging without feedback and strategy adaptation is easier during the training stage. In addition, we assume that the presence of explanations and the incremental curriculum contribute to efficient and effective human decision-making. Therefore, in \textbf{H5}, we inspect the performance of specific sorting strategies that increase in usage from sort training to sort performance test.

\section{Experiments}
\label{sec:experiments}
We first describe the implementation of a pre-test which was given to all participants at the beginning of the experiment. In addition, we report the masking and presentation of the sorting task in Section \ref{sec:materials}. In Section \ref{sec:method_design}, we explain the setup of the experiment which was carried out on a crowdsourcing platform Amazon Mechanical Turk. Results are demonstrated in Section \ref{sec:results} which is followed by a discussion of our findings.

\subsection{MaRs-IB pre-test}
\label{sec:pre_test}
We have reason to assume that the participant’s performance in the experimental task varies depending on their cognitive abilities. Recording their cognitive ability provides us with two possibilities:
\begin{enumerate}
        \item It allows statistical control of the participants' cognitive ability.
        \item It can be determined whether the mean cognitive ability in all experimental conditions is the same. If that is the case, mean differences in performance between the conditions can be interpreted without having to control for each participant's cognitive ability first.
\end{enumerate}

There have been two major suggestions for recording this control: either confront participants with a version of the experimental task itself or use a short intelligence test in the form of a Raven Matrices short test \cite{raven_matrix_test} or similar material. We argue that using the experimental task will undermine the sequential nature of our experiment, as either group would start with the low-complexity task or with the high-complexity task. A third option is for each group to start with their respective first task, which would introduce task familiarity effects (the high-complexity-first condition would have completed more high-complexity tasks than the low-complexity-first condition). This approach does not pose a good control since each group would complete a different task. 

\begin{figure}[H]
	\centering
	\includegraphics[width=0.45\textwidth]{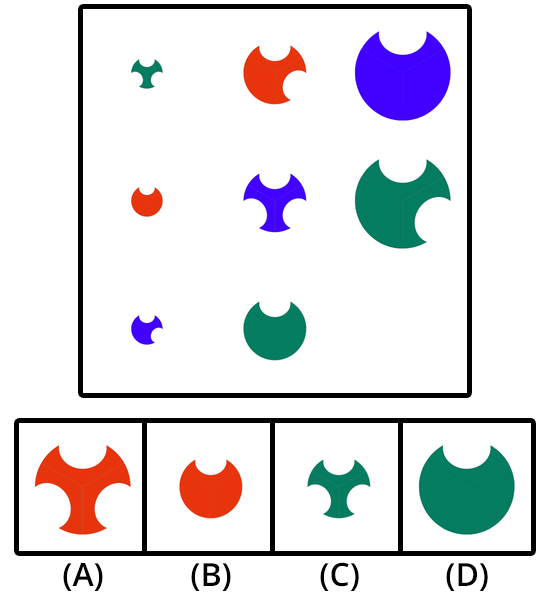}
	\caption{Each MaRs-IB item consists of an incomplete $3 \times 3$ matrix in which 8 cells with shaped icons depict a modification rule to (an) abstract shape(s). The test asks the participant to fill in the missing piece at the right bottom corner of the $3 \times 3$ matrix by choosing one of four provided options (A, B, C and D). The correct answer to this example is A.}
	\label{fig:pre_test}
\end{figure}

Using an intelligence test avoids interference with the experiment since the test should not influence strategy development in the experimental task in a way that could be considered to work against the sequential order of the experiment. However, the original Raven Matrices test and many comparable tests are expensive and cannot be digitized due to copyright. The matrix reasoning item bank (MaRs-IB) has neither of these two issues \cite{mars_ib}, thus making it an appropriate candidate for establishing a measure of intelligence in an online experiment. In addition, the authors of the MaRs-IB test set reported acceptable internal consistency and convergent validity, and reasonable test-retest reliability \cite{mars_ib}. Although there is no normalized score which would be comparable to the IQ, the MaRs-IB test can serve as a measure of the relative cognitive ability of participants. Another concern regards the open access to the material \cite{mars_ib} and familiarity with the test from other studies. This risk can be mitigated by randomised arrangement and the selection of items from the multiple test sets provided by the authors.

\subsubsection*{Application}
The authors \cite{mars_ib} provided three sets of 80 items (Figure \ref{fig:pre_test}). Each of these sets uses different symbols and is available in three further variations though colour-vision-deficient subjects were informed not to participate. The testing procedure described in the validation paper \cite{mars_ib} will be consistently completed in under 10 minutes (8 minutes test and a series of practice items beforehand) and is thus applicable for our use case. Participants will train on 10 practice items until they completed three of them correctly. An item in the full test starts with a 500 milliseconds fixation cross, followed by a 100 milliseconds white screen mask \cite{mars_ib}. The matrix will then be displayed for 30 seconds, or until the participant responded (whichever happens earlier). After 25 seconds a clock indicates that participants have 5 seconds left for this specific item \cite{mars_ib}. Participants complete these items for a total of 8 minutes. The accuracy with which the participant chooses the correct missing pattern in the given time frame is considered the measurement of pre-test performance.

\subsection{Materials} 
\label{sec:materials}
\afterpage{\begin{landscape}
\begin{figure}[h]
	\centering
	\includegraphics[width=1.5\textwidth]{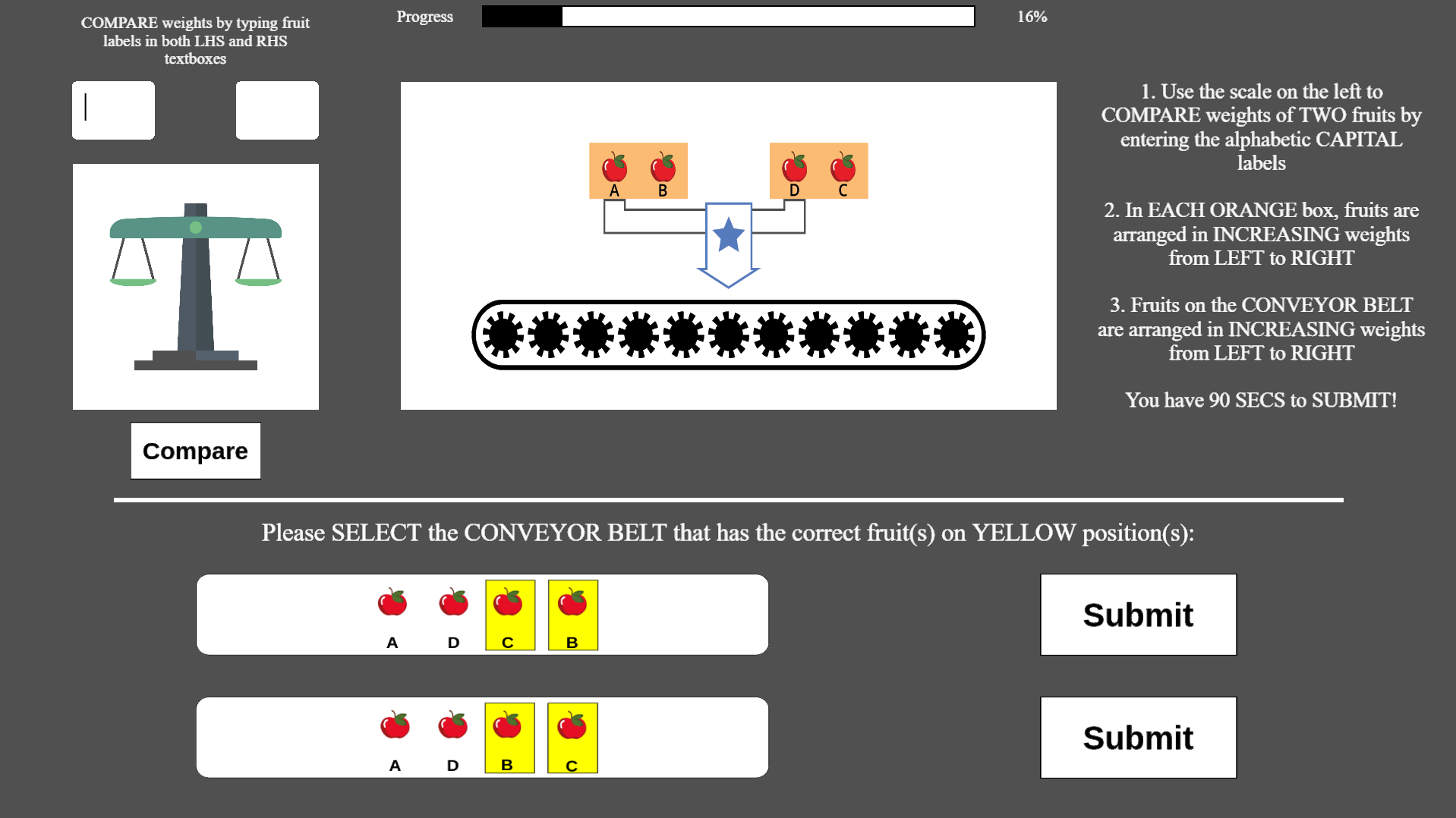}
	\caption{Participants were asked to distinguish between two alternative choices by using the provided balance scale and submit the correct sequence by pressing a button to the right.}
	\label{fig:merge_training}
\end{figure}
\end{landscape}

\begin{landscape}
\begin{figure}[h]
	\centering
	\includegraphics[width=1.5\textwidth]{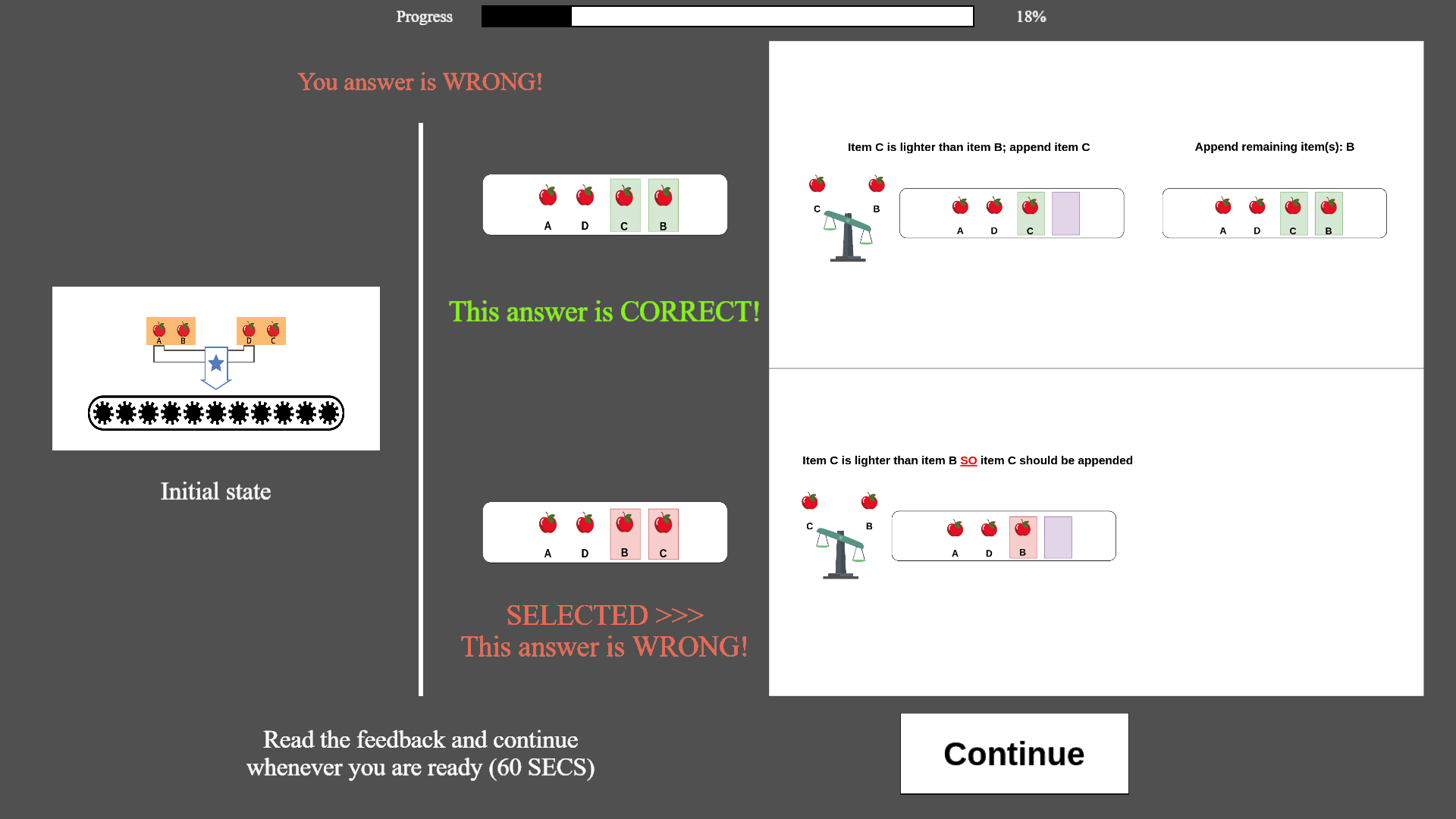}
	\caption{On the top row, the visualisations show the comparison result (fruits B and C). The textual explanations explicitly describe the objects involved (fruits B and C) and the actions performed (comparison and appending) which resulted in C being appended before B due to C's lesser weight. On the bottom row, the interface highlighted in red that fruits were out of order (B and C) and described the correct ordering. }
	\label{fig:merge_explanation}
\end{figure}
\end{landscape}

\begin{landscape}
\begin{figure}[h]
	\centering
	\includegraphics[width=1.5\textwidth]{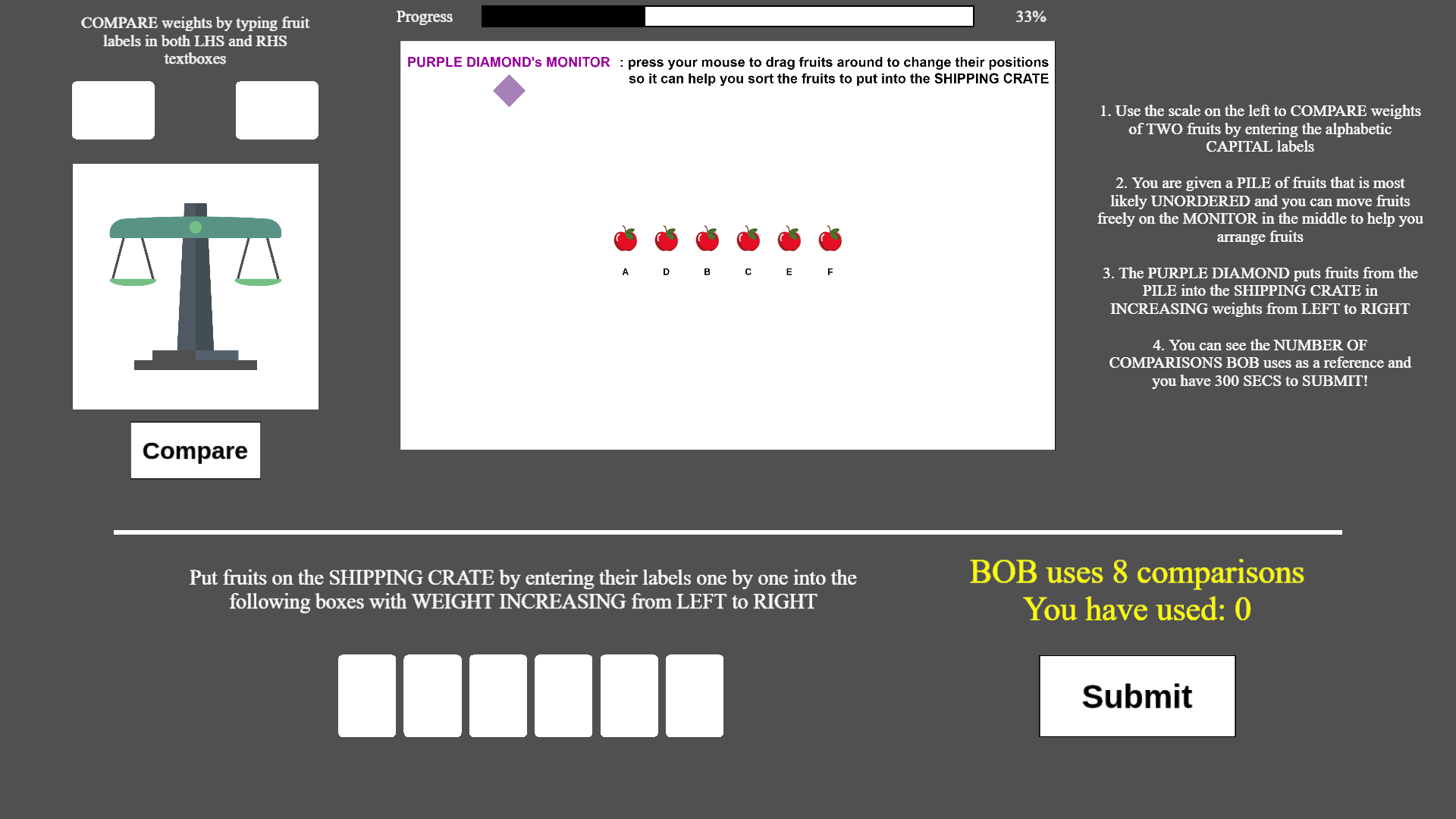}
	\caption{Sort training interface. In sort training, participants were advised to sort sequences of fruits (in this case fruits A, D, B, C, E and F) in the centre white box called the ``whiteboard'' and provided instructions to efficiently use it. The ``whiteboard'' was a device of the interface which allowed a free arrangement of the input fruit sequence by using a mouse to drag and drop fruits. The labels and fruits were intentionally randomised before the experiment so that participants would not receive the wrong impression that the sequence had been already alphabetically ordered.}
	\label{fig:sort_training}
\end{figure}
\end{landscape}}
The arrangement of objects arguably makes up a great proportion of human daily tasks. This implies a high familiarity of human participants with the task of sorting. The preferred sorting approaches of humans with no background knowledge in programming could resemble some sorting algorithms. However, we assume that learning recursive algorithms such as merge sort without a well-defined curriculum demands a great effort and applying such algorithms efficiently is challenging. 

\subsubsection*{Problem masking}

In order to prevent participants from relying on the total ordering of natural numbers, we designed an isomorphic problem that represented integer sequences by the weights of fruits in piles. Numbers were masked by visual icons of fruits and the numeric values were hidden from participants. A balance scale mechanism (Figure \ref{fig:merge_training}) was introduced to allow participants to compare the weights between two fruits and acted as an instrument for constructing the total ordering of the numbers. The immediate access to the total ordering of natural numbers was prevented and participants had to establish it themselves. This setup was used for teaching both merging and sorting. A solution of this setup corresponded to a solution of merging and sorting a sequence of numbers without changing the underlying solution structure. 

According to the definition of isomorphic problems \cite{simon1976}, the new fruit weighing problems are isomorphic to the respective problems for merging and sorting numbers. Such isomorphic translations only affect the initial stage of problem-solving while attempting to recognise a useful analogy via \textit{analogical access} but do not hinder problem-solving via \textit{analogical inference} if an analogy for the problem has been consciously recognized \cite{gentner1985,holyoak1987,reed1990}.  Owing to the difficulty of immediately recalling past experiences of merging and sorting, we anticipated that participants would initially perform less optimal merging and sorting strategies and there would be more potential for performance improvement later.

For the training and performance test of merging and sorting, we randomly sampled sequences of various lengths to provide a spectrum of problem difficulty. For each of the question sequences, we made sure that it was not trivial and required an appropriate level of computational effort for a machine program to sort. A question in the merge training and performance test involved two sub-sequences of similar sizes as inputs and the length of each sub-sequence ranged from 1 to 4. 

In the sort training and performance test, a question consists of a fruit sequence and the length of sequences varied from 6 to 10. In the pilot trials, we identified a higher usage of the insertion sort algorithm. For generating sort training and testing questions, we made an adjustment to include sequences that would lead to an advantage in the number of comparisons made when applying the merge sort algorithm over the insertion sort algorithm. This change would allow room for people to optimise their sorting strategy instead of adhering to what they already knew. The sequences in performance tests are different from those in the training stages but they share similar patterns.

\subsubsection*{Problem setup}

To make the experiment more engaging, we introduced the task background where each participant was asked to help two robots, Alice and Bob, perform tasks in two warehouses. Each robot was responsible for a distinct task and provided information about the associated task to the participant. The robot Alice was designed to teach merging while the other one Bob taught participants how to sort. To evoke an understanding of the connection between merging and sorting, at the beginning of the experiment, participants were explicitly reminded with a note of the link between task materials taught by two robots and were encouraged to pay attention to the connection. 

To learn merging, participants were asked to learn from the robot Alice and help it operate a special machine to arrange fruits on a conveyor belt. In Figure \ref{fig:merge_training}, fruits labelled with A, B, C and D were inputs to a machine called the ``blue star'' that represented the merging operation. Each question in the sections related to merging asked the participant to merge two sequences of fruits. 

The balance scale to the left of the interface page (Figure \ref{fig:merge_training}, \ref{fig:merge_explanation} and \ref{fig:sort_training}) allowed participants to compare the weights of fruits by entering two fruits' alphabetic labels on both sides of the balance. Participants were presented with instructions to use the balance scale. The balance would tilt to the side of the fruit that is heavier. A piece of text would be shown to further clarify the result of comparing two fruits. Participants were provided with the information that the two input sequences of fruits in the questions had increasing weights from left to right. Two output answer choices differed in the fruits that were highlighted with the yellow colour. Participants were asked to select from one of the two output answer choices that had the fruits in the correct order. 

Feedback was presented as shown by Figure \ref{fig:merge_explanation} on whether the participant's selection was correct. Based on the same initial problem state, this feedback provides a pair of positive and negative examples which helps contrast the correct and wrong sequence of decision-making. The white blocks on the right-hand-side of Figure \ref{fig:merge_explanation} contain textual and graphical explanations that instantiate the logic rules learned by $Metagol_O$ in Table \ref{table:merge_sort_rules}. Groups without machine-learned explanations would only see blank white boxes with no textual and graphical explanations during the merge training stage of the experiment.

\subsection{Method and design}
\label{sec:method_design}
\begin{figure}[h]
\centering
	\includegraphics[width=0.8\textwidth]{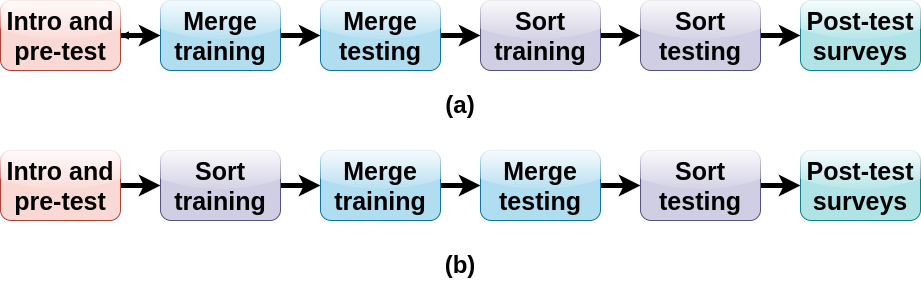}
	\caption{In setup (a), participants learned merging before learning sorting (\textbf{MS}). In setup (b), participants learned sorting before learning merging (\textbf{SM}).}
	\label{fig:experiment_design}
\end{figure}

\begin{table}[t]
\centering
\caption{Group 1 learned merging before learning sorting and was provided with visual and textual explanations of merging based on machine-learned rules. Group 2 received the same curriculum as Group 1 but without explanations. The participants in Group 3 learned sorting prior to learning merging with explanations in the same format as received by Group 1. Group 4 did not get explanations.}
	\begin{tabular}{c|c|c}
		\backslashbox{CO}{EX} & \textbf{WEX} & \textbf{WOEX}\\ \hline
		& & \\
		\textbf{MS} & Group 1 (MS / WEX) & Group 2 (MS / WOEX) \\ 
		& & \\\hline
		& & \\
		\textbf{SM} & Group 3 (SM / WEX) & Group 4 (SM / WOEX) \\ 
		& & \\
	\end{tabular}
\label{table:experiment_groups}
\vspace{-10pt}
\end{table}
The experiment\footnote{The experiment interface is available on https://github.com/LAi1997/sequential-teaching. The interface was created using PsychoPy \cite{psychopy}, an open-source package for implementing free interfaces with stimulus presentation and control in Python and JavaScript. } was implemented based on a four-group factorial $2 \times 2$ design to account for all combinations (Table \ref{table:experiment_groups}) of the two independent experimental variables, the curriculum order (\textbf{CO}) and the presentation of explanations (\textbf{EX}). Figure \ref{fig:experiment_design} shows two curriculum orders. In the curriculum shown by Figure \ref{fig:experiment_design}a, participants received training material on merging first. In the reverse curriculum demonstrated by Figure \ref{fig:experiment_design}b, subjects were trained on sorting first. 

Participants were not allowed to repeat the experiment. Every participant was assigned to one group and went through training and test with no repeat. All groups received the same material of training examples and test questions. The numbers of questions in the experiment sections merge training, merge testing, sort training and sort testing were 6, 5, 4 and 8 respectively. However, the number sequences used for merging were different from those used for sorting in order to prevent participants from reusing previous answers. 

We recorded participants' basic information such as their age, gender and education after acquiring their consent. This allows us to control the distribution of background in experiment groups. To control participants' cognitive ability, we then collected their responses in the MaRs-IB pre-test based on Section \ref{sec:pre_test}. The experiment would start with a participation consent followed by the MaRs-IB pre-test and an introduction to the task background. Participants were advised prior to the beginning of the experiment on the approximate length of the experiment and the requirements. Participants were informed that if they have colour-vision deficiency, they should not participate in the study.

In order to gain information about participants' experience relating to programming and sorting algorithms, we additionally asked participants to fill out the following questions after the completion of all performance tests: 1) whether participants have a degree in computer science or a certificate in programming, 2) if they have studied or are learning sorting algorithms in school and 3) whether they know or have applied any sorting algorithms while being presented with a list of sorting algorithms (two of which are distracting choices and do not exist). 

Task-related information was presented via the interface throughout the experiment. All groups received a performance test of sorting last. This experiment design ensured all groups experienced the same amount of training and performance tests respecting the curriculum order. 

In merge training, participants first answered the displayed question illustrated by Figure \ref{fig:merge_training} and were taken to the next page where feedback and explanations were presented. In Figure \ref{fig:merge_explanation}, explanations are presented in two rows which demonstrate the correct and the wrong action sequences. The format in which participants were asked to input their responses is the same in the merge performance test, sort training and sort performance test. When providing a sequence response, participants were required to input labels representing the fruit sequence that they believed to have increasing weights from left to right. Another robot instructor Bob helped participants learn to operate the machine ``purple diamond'' to sort fruits. 

In the sort training and performance test, participants should put fruits into shipping boxes by entering fruit labels into a sequence of text boxes. In the sort training (Figure \ref{fig:sort_training}), participants were informed of the number of comparisons their robot instructor Bob made to sort the input sequences. Participants were asked to sort the same input sequences and encouraged to use fewer comparisons. The intention of this competition was to provide an incentive for participants to consciously think about how to revise their strategy in order to improve sorting efficiency. This information was accessible by participants in all groups during the sort training. In the sorting sections, participants were also provided with the balance scale instrument.

Participants were encouraged to take two sessions of one-minute rest, one after the MaRs-IB pre-test and one prior to the final performance test of sorting. The break sessions allowed participants to recharge and regain attention to perform tasks in later stages. The progress bar at the top of the interface showed the percentage of materials that the participants had completed with respect to all materials in the experiment. Participants could use the progress bar to estimate the time that they would need to finish the experiment within the allocated session. For all training and performance test sections, we recorded response time, answers and the comparisons made for analysis. The final section included four reflective questions. We asked participants whether they understood the connection between the two tasks or used what they learned in one task for the other task. We also checked their background in programming and sorting algorithms in the survey at the end of the experiment. 

\subsection{Results}
\label{sec:results}
\begin{figure}[t]
    \centering
    \begin{tabular}{cc}
        \begin{subfigure}{0.48\textwidth}
	        \includegraphics[width=\linewidth]{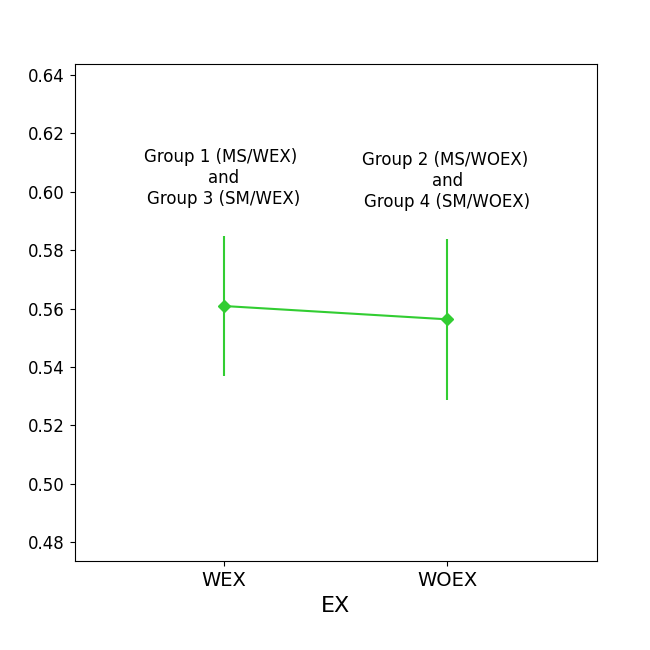}	
	        \caption{Mean \textbf{merge} performance test scores and standard errors.}
	        \label{fig:merge_performance_score}
        \end{subfigure} &
        \begin{subfigure}{0.48\textwidth}
	        \includegraphics[width=\linewidth]{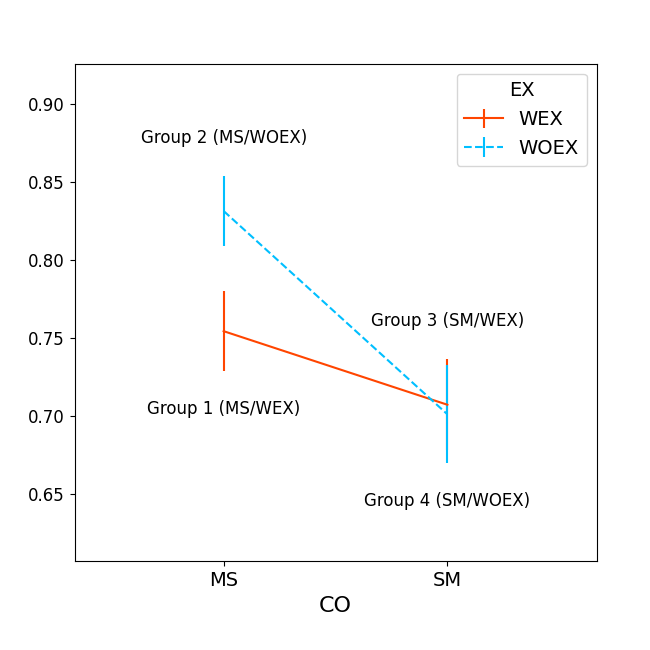}
	        \caption{Mean \textbf{sort} performance test scores and standard errors.}
	    \label{fig:sort_performance_score}
        \end{subfigure}
    \end{tabular}
    \caption{Performance test scores (\textbf{PS}) of four groups, Group 1 (\textbf{MS}/\textbf{WEX}) Group 2 (\textbf{MS}/\textbf{WOEX}), Group 3 (\textbf{SM}/\textbf{WEX}) and Group 4 (\textbf{SM}/\textbf{WOEX}) with standard errors. Across the x-axis in (a), data points indicate the change in the mean merge performance score as a result of learning with explanations. In (b), the red line shows the mean sort performance score when participants learned with explanations and the blue line represents the mean sort performance score when participants learned without explanations. Across the x-axis in (b), data points show the change in the mean sort performance score as a result of altering the curriculum arrangement. }
    \label{fig:performance_score}
\end{figure}
Prior to the experiment, we conducted two experimental trials with one sample of students from the University of Bamberg and another student sample from Imperial College London. The trial studies provided additional insights into the interface and helped us revise the experimental design. However, due to the limited scale of the trials, we were not able to observe any meaningful outcome. We proceeded to recruit a larger sample of 124 mixed-background participants from Amazon Mechanical Turk (AMT). These 124 participants from AMT were randomised and assigned to the four groups. A partition of 79 participants was created based on the MaRs-IB test and those included had a test accuracy within one standard deviation ($\sigma = .169$) of the mean ($\mu = .654$). Demographically, this partition had a lower male-to-female ratio (36 to 42, one opted out), most participants were between 25 to 54 in age (61 of 79) and almost all had a college degree or higher (67 of 79). 

\subsubsection*{Analysis of task performance}

We focused on this AMT sample partition and employed a quantitative approach to test hypotheses \textbf{H1} to \textbf{H5}. The AMT partition had an even distribution of participants in the four groups, Group 1 (\textbf{MS} and \textbf{WEX}, $n = 20$), Group 2 (\textbf{MS} and \textbf{WOEX}, $n = 20$), Group 3 (\textbf{SM} and \textbf{WEX}, $n = 20$) and Group 4 (\textbf{SM} and \textbf{WOEX}, $n = 19$). Figure \ref{fig:merge_performance_score} and Figure \ref{fig:sort_performance_score} present the mean performance score (\textbf{PS}) of the groups in the merge performance test and sort performance test. For each performance test, we carried out an ANOVA test to assess the effect of the independent experimental variables followed by a Tukey’s HSD test. We used a significance criterion $\alpha = .05$ for both tests. 

Figure \ref{fig:merge_performance_score} illustrates almost no change to the mean merge performance score when explanations were introduced. The one-way ANOVA test did not demonstrate a significant effect of the machine-learned theory (\textbf{EX}) on the merge performance test score (\textbf{PS}). Figure \ref{fig:sort_performance_score} illustrates a pattern that switching from the curriculum order \textbf{MS} to \textbf{SM} results in a reduction in performance. The two-way ANOVA test indicated a significant effect ($F=10.4, p = .001 < .05$) of curriculum order (\textbf{CO}) on sort performance test score (\textbf{PS}). The post-hoc Tukey's HSD test confirmed a significantly beneficial effect ($p=.001 < .05$) of the sequential teaching curriculum \textbf{MS} (Group 1 and Group 2) over \textbf{SM} (Group 3 and Group 4). The two-way ANOVA test did not show a significant effect of the machine-learned theory (\textbf{EX}) on sort performance test score (\textbf{PS}) and there is no evidence of an interaction effect between curriculum order (\textbf{CO}) and presence of explanations (\textbf{EX}).  

\begin{table}[t]
\centering
\caption{Mean strategy frequency in sort training and sort performance test. Each human sorting trace in the sort training and sort performance test was matched to one of the following categories of sorting algorithms, bubble sort ($BS$), dictionary sort ($DS$), insertion sort ($IS$), merge sort ($MS$), quick sort ($QS$), a combination of insertion and dictionary sort ($Hybrid$) and unclassified ($Other$). The number of human sorting traces in each category was averaged by the total number of traces. We highlight positive differences greater than .05 in grey and in bold font. In each group,  the highlighted difference is the largest increase in algorithm application frequency from sort training to sort performance test. }
	\begin{tabular}{c|c|c|c|c|c|c|c}
		\backslashbox{PS}{Categories} & $BS$ & $DS$ & $IS$ & $MS$ & $QS$ & $Hybrid$ & $Other$ \\ \hline
		\textit{Group 1 (MS/WEX)} & -- & -- & -- & -- & -- & -- & --\\ \hline
		Training & .012 & .075 & .150 & .000 & .175 & .162 & .425\\ \hline
		Performance test &  .056 & .094 & .162 & .025 & .238 & .175 & .250 \\ \hline
		Differences & .044 & .019 & .012 & .025 & \cellcolor{lightgray} \textbf{.063} & .013 & -.175 \\ \hline
		\textit{Group 2 (MS/WOEX)} & -- & -- & -- & -- & -- & -- & -- \\ \hline
		Training & .000 & .062 & .162 & .025 & .162 & .225 & .362 \\ \hline
		Performance test & .012 & .038 & .181 & .100 & .194 & .181 & .294 \\ \hline 
		Differences & .012 & -.024 & .019 & \cellcolor{lightgray}\textbf{.075} & .032 & -.044 & -.068\\ \hline
		\textit{Group 3 (SM/WEX)} & -- & -- & -- & -- & -- & -- & -- \\ \hline
		Training & .012 & .050 & .088 & .038 & .225 & .175 & .412 \\ \hline
		Performance test & .019 & .138 & .100 & .025 & .244 & .119 & .356\\ \hline 
		Differences & .007 & \cellcolor{lightgray}\textbf{.088} & .012 & -.013 & .019 & -.056 & -.056 \\ \hline
		\textit{Group 4 (SM/WOEX)} & -- & -- & -- & -- & -- & -- & -- \\ \hline
		Training & .000 & .079 & .184 & .026 & .158 & .237 & .316\\ \hline
		Performance test & .013 & .099 & .243 & .053 & .158 & .237 & .197\\ \hline
		Differences & .013 & .020 & \cellcolor{lightgray}\textbf{.059} & .027 & .000 & .000 & -.119
	\end{tabular}
\label{table:sorting_strategy} 
\end{table}

\subsubsection*{Analysis of human sorting strategy}
In addition to the performance score, we analyse the correspondence between human sorting strategies and machine sorting algorithms. For sorting, sequential decision-making is revealed through the actions that human sorters make to arrange items according to a specified order. In particular, we consider the comparisons of objects as traces of decision making which indicate the sorting strategy used by human sorters and the sorting algorithm implemented by a machine program.



When there is a consistently high correlation of comparison execution between a human sorter and a machine algorithm, we assume that this human executes a strategy that closely implements the machine algorithm. Considering the target human population has limited programming experience, we additionally assume that a human sorting strategy would implement at best quick sort, merge sort, insertion sort, bubble sort or dictionary sort. Based on the results and preferences of the trial participants, we include hybrid variants of insertion sort and dictionary sort. These hybrid variants initially perform insertion sort until the intermediate result reaches length k and then execute binary searches of the conventional dictionary sort to insert the rest of the numbers into the partially sorted sequence. In total, we implemented twenty-four algorithms including the conventional implementations of quick sort, merge sort, insertion sort, bubble sort and dictionary sort as well as their variants (different pivoting, merge and insertion implementations). 

\begin{example} Given $Ns = $ [4, 6, 5, 2, 3, 1] is a sequence of numbers for a human participant $h$ and a set of machine algorithms to sort. Let $M_h$ denote the human sorting algorithm and $ht(Ns, M_h)$ = [(6, 4), (5, 2), (3, 1), (4, 2), (5, 4), (6, 5), (2, 1), (3, 2), (4, 3)]. We are able to find a close match of the human trace. The machine trace is produced by the bottom-up variant of the merge sort algorithm denoted as $M'$ and $mt(Ns, M')$ = [(4, 6), (5, 2), (2, 4), (4, 5), (5, 6), (3, 1), (1, 2), (2, 3), (3, 4)]. For each of the $N \times N$ combinations of number pairs, we check whether it is a member of the traces. Two symmetric pairs of numbers, e.g. (4, 6) and (6, 4), are considered identical. We create the following $2 \times 2$ contingency table. 
\begin{table}[h]
    \centering
    \begin{tabular}{c|c|c}
		& Not in human trace & In human trace \\ \hline
		Not in machine trace & 13 & 1 \\ \hline
		In machine trace & 1 & 10 \\
\end{tabular}
\end{table}
The Chi-squared test\footnote{We add one to all frequency counts to avoid including zeros in the table for the Chi-squared test. } in this example has a $X^2=14.3$ with $p < .001$ which confirms the correlation of number pairs in $ht(Ns, Mh)$ and $mt(Ns, M')$. For common pairs of numbers in $ht(Ns, Mh)$ and $mt(Ns, M')$, we obtain their ranks and compute Spearman's rank correlation coefficient. The correlation coefficient is .9 and $p < .001$ which confirms the monotonic relationship between the two traces. 
\label{ex:trace_analysis}
\end{example}

Given an input set of integers $Ns$ and a sorting algorithm $M$, a sorting trace $M(Ns)$ is a list of pairs $<n_i, n_j>$ denoting the comparisons of integers made by $M$ where $n_i, n_j \in Ns$. We employ two statistical tests (Example \ref{ex:trace_analysis}) to estimate the correspondence between a human sorting trace and a machine sorting trace. Let $n$ be an integer and $Ns$ denote the set of integers in the interval $[1, n]$. Let $Tr$ be a cross product defined by $Tr: Ns \times Ns$, a comparison in a trace is a pair $<n_i, n_j> \in Tr$ where $n_i, n_j \in Ns$. Given a human trace with sorting algorithm $M_h$ and a machine trace with algorithm $M$, for every pair $<n_i, n_j> \in Tr$, we check if $<n_i, n_j>$ is a member of $ht(Ns, M_h)$ and if $<n_i, n_j>$ is a member of $mt(Ns, M)$. We then perform a Chi-squared test of independence with a significance level $\alpha = .025$ using the $2 \times 2$ contingency table and check for rejection. The significance criterion of the Chi-squared test is set at .025 to reduce the probability of mismatching a human trace with a wrong machine trace. A rejection of the Chi-squared test's null hypothesis implies an association between comparisons made by the human sorter and comparisons made by the program. Once a test is rejected,  we exclude comparison pairs that are not in the intersection of $ht(Ns, M_h)$ and $mt(Ns, M)$ from both traces and compute Spearman's rank correlation coefficient \cite{spearman_rank}. The intention of computing Spearman's rank correlation coefficient is to examine the ordering alignment of compared integer pairs. 

Since performing comparisons in the reverse order of the machine trace does not correspond to a reverse execution of the algorithm, we confirm a match ($\alpha = .05$) between a human strategy and a machine algorithm only when Spearman's rank correlation coefficient is positive. 

Table \ref{table:sorting_strategy} demonstrates the change in the distribution of strategy as the difference in mean strategy frequency from the training and performance test. In addition, Table \ref{table:sorting_strategy_contingency} shows the strategy adaptation of participants. Participants were categorised with respect to whether they had applied a particular strategy in the training phase and in the performance test. We then employ McNemar's tests (approximated by binomial distribution for small frequencies) to determine if there is a difference in the number of participants who applied the strategy in the test and who used it in training. For strategies that have been identified to have increased applications from the training to performance test, we additionally examine the performance of these strategies using t-tests. 

\begin{table}[t]
\centering
\caption{The number of participants in Group 1 to 4 who had applied one of the highlighted algorithms (QS, MS, DS and IS) in the training and the performance test.  In Table \ref{table:sorting_strategy}, algorithms QS, MS, DS and IS have increased application frequency in Group 1, 2, 3 and 4 respectively. Every participant was categorised with respect to whether they had applied the strategy at least once in sort training and performance test. While apart from those four algorithms analyses were performed in each group with other strategies, the results presented were the most distinguished. }
	\begin{tabular}{c c c c c}
		\textit{Group 1 (MS/WEX)} &  &  & Training & \\
		\textbf{Quick Sort} & & Application & No application & Total \\ \hline
		& Application & 7 & 6 & 13\\ \cline{2-5}
		Performance test & No application & 0 & 7 & 7 \\ \cline{2-5}
		& Total & 7 & 13 & 20 \\ \hline
		\\ 
		\textit{Group 2 (MS/WOEX)} &  &  & Training &\\
		\textbf{Merge Sort} & & Application & No application & Total \\ \hline
		& Application & 2 & 7 & 9\\ \cline{2-5}
		Performance test & No application & 0 & 11 & 11\\ \cline{2-5}
		& Total & 2 & 18 & 20 \\ \hline
		\\ 
		\textit{Group 3 (SM/WEX)} &  &  & Training & \\
		\textbf{Dictionary Sort} & & Application & No application & Total \\ \hline
		& Application & 3 & 7 & 10 \\ \cline{2-5}
		Performance test & No application & 1 & 9 & 10 \\ \cline{2-5}
		& Total & 4 & 16 & 20 \\ \hline
		\\ 
		\textit{Group 4 (SM/WOEX)} &  &  & Training &\\
		\textbf{Insertion Sort} & & Application & No application & Total \\ \hline
		& Application & 7 & 11 & 18\\ \cline{2-5}
		Performance test & No application & 0 & 1 & 1\\ \cline{2-5}
		& Total & 7 & 12 & 19 \\
	\end{tabular}
\label{table:sorting_strategy_contingency} 
\vspace{-10pt}
\end{table}

Group 1 (\textbf{MS}/\textbf{WEX}) and Group 3 (\textbf{SM}/\textbf{WEX}) both received explanations (\textbf{WEX}). According to Table \ref{table:sorting_strategy}, from training to performance test the average use of quick sort in Group 1 increased by .063 and the average application of dictionary sort in Group 3 improved by .088. McNemar's test ($\alpha = .05$) reports a significant change in the application of quick sort in Group 1 ($p = .031 < .05$). There is no significant difference in the application of dictionary sort in Group 3. Based on Table \ref{table:sorting_strategy_contingency}, in Group 1, we can observe that more participants used quick sort in the performance test (13) than in the training stage (7). 

Further t-tests ($\alpha = .05$) on performance test scores showed that in Group 1 quick sort like approaches ($\mu=.834, \sigma=.274$) achieved a higher mean score ($p = .043 < .05$) than the rest of the strategies ($\mu=.729, \sigma=.337$) and in Group 3 responses corresponding to dictionary sort like approaches ($\mu=.885, \sigma=.288$) had a better performance ($p = .007 < .05$) than the other strategy categories ($\mu=.679, \sigma=.373$). 

Group 2 (\textbf{MS}/\textbf{WOEX}) and Group 4 (\textbf{SM}/\textbf{WOEX}) did not receive explanations (\textbf{WOEX}). From Table \ref{table:sorting_strategy}, in Group 2 the frequency of application of merge sort increased by .075 and in Group 4 the average use of insertion sort increased by .059. McNemar's tests show significant changes in the application of merge sort in Group 2 ($p = .016 < .05$) and insertion sort in Group 4 ($p = .001 < .05$). In Table \ref{table:sorting_strategy_contingency}, we can observe that in Group 2 more participants were using merge sort in test (9) than in training (2). In Group 4, almost all participants used insertion sort in the performance test (18) compared with in the training (7).

Additional t-tests ($\alpha = .05$) showed no performance difference in Group 2 between merge sort like approaches ($\mu=.872, \sigma=.209$) and the other strategies ($\mu=.827, \sigma=.288$). In contrast, insertion like approaches ($\mu=.877, \sigma=.285$) in Group 4 achieved a higher mean score ($p = .001 < .05$) than the other strategies ($\mu=.644, \sigma=.399$).

\subsection{Discussion}
\label{sec:discussion}
In Table \ref{table:hypothesis_confirmation}, we present experimental hypotheses \textbf{H1} to \textbf{H5} and summarise their test outcomes. Since this work investigates the effect of curriculum order, we have acknowledged the possibility of recency and primacy effects on human sorting performance. Recency and primacy effects \cite{murdock1962} are cognitive biases which lead to better recall of items and concepts that are positioned towards the beginning and the end of a sequence. As it is part of our experimental plan to explore how human comprehension is affected by different sequential orders of concepts, it is difficult to completely eliminate the impact of position effects. In addition, Table \ref{table:sorting_strategy_contingency} shows that various degrees of adaptations to efficient sorting algorithms happened in multiple groups. This indicates the adjustment of participants to different decision-making processes beyond making better or worse recall as a result of learning in sequential curricula. 

Based on the definition of $sorter/2$ learned by $Metagol_O$ after learning $merger/2$ in Table \ref{table:merge_sort_rules}, the program size is $u=3$ and the number of predicates including those used in $merger/2$ is $p=8$. For the definition of $sorter/2$ learned by $Metagol_O$ without learning $merger/2$ in the same table, the size of $sorter/2$ is $u+k = 5$ and the number of predicates used is $p+c = 6$. According to the Conjecture \ref{conjecture:sequential_teaching_improvement}, $3 \cdot\ln (8) < 5 \cdot\ln (6)$ and learning merging before learning sorting results in a reduction in the size of the hypothesis space associated with the target hypothesis of sorting. The ANOVA test and Tukey's HSD test on sort performance test scores (Figure \ref{fig:sort_performance_score}) demonstrated a beneficial effect from curriculum order. A higher sorting performance has been observed only in groups with the incremental curriculum setup. Since all four groups received the same material in each stage of the experiment so we can eliminate the possibility of learning merging or sorting being a major effect which led to the observed difference in sorting performance. Hence, we attribute the improvement in the performance score of sorting to the incremental curriculum order. This evidence confirms hypothesis \textbf{H1} and $E_{seq}$($C_{MS/WEX}$, $C_{SM/WEX}, sorter$) + $E_{seq}$($C_{MS/WOEX}$, $C_{SM/WOEX}, sorter$) > 0 which supports Conjecture \ref{conjecture:sequential_teaching_improvement}.

\begin{table}[t]
	\centering
	\caption{\textbf{H1} to \textbf{H3} are hypotheses concerning the effects of curriculum order and explanations on human sorting comprehension. Hypotheses \textbf{H4} and \textbf{H5} relate to explanatory effects on human comprehension of merging and sorting strategy. At the top of table, H stands for \textbf{h}ypothesis, T denotes \textbf{t}est outcome. In the rightmost column, C is an abbreviation for \textbf{c}onfirmed, and N stands for \textbf{n}ot confirmed.}
	\begin{tabular}{ p{0.3cm} p{10cm} p{0.25cm} }
		H &  & T \\ \hline 
		\\ [-0.5em]
		\textbf{H1} & Learning merging before learning sorting leads to a beneficial effect on human comprehension of sorting with respect to learning sorting before learning merging & C \\
		& & \\
		\textbf{H2} & Learning merging with explanations results in a beneficial explanatory effect on human comprehension of sorting with respect to learning merging without explanations & N \\
		& & \\
		\textbf{H3} & Learning merging with explanations further increases the beneficial effect of curriculum order on human comprehension of sorting with respect to learning merging without explanations & N \\
		& & \\
		\hline
		& & \\
		\textbf{H4} & Learning merging with explanations generated from rules without a low cognitive cost does not result in a beneficial explanatory effect on human comprehension of merging & C \\
		& & \\
		\textbf{H5} & Learning merging with explanations before learning sorting leads to adaptation of efficient human sorting strategies with better performance & C \\
	\end{tabular}
	\label{table:hypothesis_confirmation}
\end{table}

Results on merge and sort performance tests did not show explanatory effects of the $merger/2$ rules learned by $Metagol_O$. The ANOVA test did not show a significant effect of explanations on human comprehension and there was no significant interaction effect on human comprehension. Therefore, we reject hypotheses \textbf{H2} and \textbf{H3} due to the lack of evidence. Learning merging with explanations did not improve or degrade human comprehension of sorting in regard to learning merging without explanations. In addition, the ANOVA test on the merge performance test scores did not demonstrate an effect of explanations on human comprehension. This result confirms \textbf{H4} and supports Conjecture \ref{conjecture:cognitive_window}. 

Since merging was taught in isolation independent from other secondary or tertiary concepts, we refer to the cognitive window (Remark \ref{remark:merge_beneficiality}) for a plausible account of this lack of explanatory effect on human comprehension of merging. The cognitive cost of executing explained rules of merging is not sufficiently lower than the cognitive cost of operating a merging solution without the auxiliary information. Owing to limited cognitive capacities, a reduction in computational cost usually corresponds to an improvement in performance. However, solutions of merging after receiving explanations were not sufficiently optimised to yield an observably higher performance score compared with human primitive solutions in the absence of explanations. Given the tasks and the sequential teaching setup, an incremental curriculum order had a significant effect which improved human comprehension. The results provided a demonstration of the beneficial effect of sequential teaching on human comprehension.

Furthermore, we examined human comparisons and estimated sorting algorithms that best resembled human sorting traces. This analysis (Table \ref{table:sorting_strategy}) led to the recognition of four possible sorting strategy adaptations. From Table \ref{table:sorting_strategy_contingency} and McNemar's tests, we show that there is a significant difference in the algorithm application of quick sort in Group 1, merge sort in Group 2 and insertion sort in Group 4. We argue that these results support a higher proportion of strategy adaptations to these algorithms. The additional t-tests compared the sorting performance of adapted strategies with other strategies. Quick sort in Group 1, dictionary sort in Group 3 and insertion sort in Group 4 had higher performance in comparison with other algorithms. Table \ref{table:strategy_adaptation_summary} summarises the participants' adaptations to algorithms and the performance of responses with these adaptations. 

\begin{table}[t]
	\centering
        \caption{Summary of the significance of participant's strategy adaptation and performance improvement of the adapted algorithm compared with other algorithms. A tick denotes a significant analysis result, and a cross indicates an insignificant analysis outcome. }
        \begin{tabular}{ c | c | c | c }
         & Algorithm & Is adaptation & Is performance \\
        Group & adapted & significant & improvement significant \\ 
        & & & \\ \hline
        & & & \\ 
        Group 1 (\textbf{MS}/\textbf{WEX}) & QS & \checkmark & \checkmark \\ 
        & & & \\ \hline
        & & & \\ 
        Group 2 (\textbf{MS}/\textbf{WOEX}) & MS & \checkmark & \textit{X} \\ 
        & & & \\ \hline
        & & & \\ 
        Group 3 (\textbf{SM}/\textbf{WEX}) & DS & \textit{X} & \checkmark \\ 
        & & & \\ \hline
        & & & \\ 
        Group 4 (\textbf{SM}/\textbf{WOEX}) & IS & \checkmark & \checkmark \\ 
        & & & \\ 
        \end{tabular}
        \label{table:strategy_adaptation_summary}
\end{table}

Strategy adaptations in Group 1 (\textbf{MS}/\textbf{WEX}) and Group 4 (\textbf{SM}/\textbf{WOEX}) were significant and resulted in higher performance scores of the associated responses compared with the other strategies. After sort training, Group 1 used a higher volume of quick sort like strategies. The quick sort algorithm is computationally efficient with an average linearithmic run-time and a worst-case quadratic run-time. Utilising the efficiency benefit of concentrating on parts of the original problem one at a time, quick sort creates a pivot to construct two sorted sub-sequences. Based on the follow-up t-tests, adaptation to quick sort led to better comprehension compared with responses that used the other sorting strategies. Since in Group 1 participants learned merging with explanations before learning sorting, this refutes the null hypothesis of \textbf{H5}. This positive difference in sorting performance from using quick sort in Group 1 allows us to confirm hypothesis \textbf{H5}. 

The t-test results of Group 1 and 3 (summarised in Table \ref{table:strategy_adaptation_summary}) suggest that curriculum order does not affect the performance of adapted strategies since Group 1 experienced the incremental curriculum and Group 3 received the decremental curriculum. This observation contrasts with the overall performance analysis which shows curriculum order affects the performance of sorting. It implies that the decremental curriculum in Group 3 reduces the performance of strategies that have a lower change in application rate. We postulate that without the benefit of the incremental curriculum and a compact hypothesis space (Conjecture \ref{conjecture:sequential_teaching_improvement}), it is difficult to develop rules that consistently represent the other algorithms (quick sort and merge sort). In addition, we observed in Group 1 that learning incrementally with explanations led to a significant adaptation of an efficient sorting strategy (quick sort) which also has enhanced performance. This shows the association of a higher task competency with the development of novel problem solutions in novices. 

\subsubsection*{Impacts of explanations on performance}

As summarised in Table \ref{table:strategy_adaptation_summary}, in Group 1 and 3 where explanations were presented, the adaptations to efficient sorting approaches correspond to improved performance. A significant number of participants adapted to quick sort like approaches and some participants adjusted to dictionary sort like strategies in Group 3. Although the dictionary sort algorithm shares many similar traits to insertion sort, dictionary sort is considered a divide-and-conquer algorithm which iteratively makes binary searches for the correct position of an object in a sequence. 

We attribute this phenomenon to a better understanding of efficiently merging: two input sequences are sorted and redundant comparisons can be avoided by interleaving comparisons of fruits from input sequences. Provided explanations emphasised the optimal problem-solution structure and illustrated the action sequence by walking through examples with participants. The contexts provided by explanations and examples are effective for the human learning of abstract concepts \cite{aleven2002effective,anderson1997role,berry1995implicit,reed1991use}. We suggest that in the presented teaching setup, explanations of merging involving examples grounded the concept of binary selection and contextualised the construction of problem solutions that utilised this information. As a result, problem solutions that incorporated this idea had less potential for errors and achieved higher performance. This proposition can be partially supported by the increase in the usage of insertion sort like strategies in Group 4 (\textbf{SM}/\textbf{WOEX}) where explanations were absent. While insertion sort like strategies correlated to better scores in sort performance test responses of Group 4, the insertion sort algorithm does not involve a divide-and-conquer or binary selection component to sorting and does not share these features with quick sort or dictionary sort.

\subsubsection*{Impacts of curriculum order on strategy adaptation}

Another observation from Table \ref{table:sorting_strategy_contingency} and \ref{table:strategy_adaptation_summary} is that a significant number of responses in Group 2 (\textbf{MS}/\textbf{WOEX}) adapted to merge sort. While this strategy adaptation of Group 2 to merge sort did not correlate with a higher performance score compared with the other strategies, divide-and-conquer algorithms such as quick sort and merge sort that require an advanced understanding of computational algorithms could be rather difficult for novices to comprehend in a limited time frame. Based on Table \ref{table:sorting_strategy_contingency} and \ref{table:strategy_adaptation_summary}, it is noteworthy that a significant number of participants in Group 1 and 2 appeared to have developed sorting approaches similar to these well-known divide-and-conquer algorithms when they were presented with concepts with increasing complexity. A tentative account of this phenomenon can be referred to Conjecture \ref{conjecture:sequential_teaching_improvement}. In Conjecture \ref{conjecture:sequential_teaching_improvement}, two curricula are compared in terms of the size of the composite hypothesis spaces of learning some target predicates. Learning concepts in an incremental fashion reduces the size of the hypothesis space. Since humans have limited working memory capacity, it becomes easier for humans to find a consistent hypothesis that incorporates new information in the associated hypothesis space. 

Despite performance improvements of adapted strategies in both Group 1 and 3, the adaptation of an efficient strategy was significant and led to better performance in only Group 1, as summarised in Table \ref{table:strategy_adaptation_summary}. As Group 1 exclusively experienced incremental curriculum and machine-learned explanations, this result highlights the impact of combining both components on human learning for promoting task performance and developing novel problem solutions. Although it cannot be assessed by the present framework whether the participants mentally formulated quick sort or merge sort in an explicit way, further investigations of sequential teaching could devise machine learning to derive human strategy from behavioural traces.

\section{Conclusions and further work}
\label{sec:conclusion}

Previous publications \cite{lun2021,US2018} on the topic of comprehensibility have investigated the classification of explanatory machine learning into beneficial and harmful categories. The current work proposes an extension of frameworks of comprehensibility \cite{US2018} and explanatory effects \cite{lun2021} to account for the effects of sequence teaching. Owing to the reduction in the size of the hypothesis space \cite{playgol}, we hypothesise that learning concepts with increasing complexity enables humans to learn the target hypotheses given limited working memory capacity. This conjecture is supported by our empirical results. 

In the limited scope of our experiment which focuses on human learning of sorting algorithms, we have identified an instance of sequential teaching curricula in which learning merging before learning sorting results in a better human comprehension of sorting in contrast to learning sorting before learning merging. This result demonstrates an improvement in human comprehension as a result of learning concepts with increasing complexity. We refer to the cognitive window \cite{lun2021} to account for the lack of explanatory effects on human comprehension of merging. It is difficult to improve human comprehension from explanations generated from machine-learned rules when the cognitive cost of the task is already low. In this case, explanations do not cause a sufficiently high improvement in the cognitive cost of performing the task. However, we have taken a rather conservative approach which has led to interesting preliminary results of sequential teaching in a specific domain. While this work is an extended investigation of machine learning comprehensibility based on translated logic programs, we also acknowledge the potential of developing comprehensibility definitions beyond symbolic machine learning and our noise-free framework for future work. 

Instruction-based teaching approaches have the advantage of clarity and directness for human learning \cite{sweller2007} and allow previous knowledge to be integrated with new problem-solving contexts \cite{mayer2004}. However, such teaching approaches may impose an over-restriction on the links between actions and outcomes and therefore fail to trigger an individualised understanding generalised from the material \cite{bruner1961}. In the human trial, we employed a mixed approach to teaching that combined instruction-based learning with discovery learning. We explored a machine-human teaching interaction where curricula allow a higher degree of freedom for learners to re-discover computationally efficient algorithms. We observed adaptations of human sorting strategies to utilise advanced techniques of divide-and-conquer. Such strategy adaptations were observed in humans who learned with explanations and in humans who learned concepts with incremental complexity. When both conditions were fulfilled, humans adapted to an efficient divide-and-conquer strategy with higher performance. We attribute these results to the contextualisation of abstract concepts by explanations and the accessibility of learned knowledge from the reduction in complexity in searching the hypothesis space. In these cases, human learners were able to adapt sorting strategies reminiscent of efficient machine sorting algorithms. Although the analysed participant sample had no background in programming, they adapted sorting strategies by employing divide-and-conquer techniques commonly found in efficient computational sorting algorithms. While not all inspected human learners managed to consistently perform these sorting algorithms after studying from short learning sessions, being able to develop ``near-miss'' versions of the merge sort algorithm is a remarkable achievement for those with no background in programming. 

An exciting prospect of sequential teaching curricula is that machines and humans could take up more active roles to enable two-way cooperative learning \cite{human_robot_interactions}. Sequential teaching curricula can be extended to represent the coordination between humans and machines in a back-and-forth fashion.  Human implicit decision-making (System 1) could be made explicit (System 2) \cite{kahneman2011thinking} to benefit human problem-solving \cite{Schmid11} by machine-learned theories via a process known as behavioural cloning \cite{BRATKO1995143}. Traces of human problem solutions can be provided as inputs to an ILP system, which might produce explicit algorithms from the traces to present to the human as explanations. A ``clean up'' effect on a human's behaviours can be achieved based on the error estimation of the machine-learned human algorithm in training. The debugging of behaviours guided by the prediction of error of a human strategy can be beneficial in intelligent tutoring \cite{ZellerSchmid16} and for cooperative programming in algorithmic debugging \cite{shapiro1982algorithmic}. 

Future work may also explore the phenomenon of human re-discovery of well-established domain knowledge. An investigation of partially guided development of efficient problem-solving strategies would require exposing the implicit cognitive processes. External tools such as eye-tracking devices \cite{eye_tracking} can be used to capture subtle motions of the human body to help us infer implicit human decision-making. Our observations have shown the creativeness of human learning in exploiting the analogical transfer of partial solution structure. For machine learners, insights obtained in other domains can be utilised to learn flexible and efficient problem solutions \cite{bratko_concept_discovery,bratko_robot_discovery_ILP}. For human learners, the division of information into smaller ``chunks'' and the presentation of rule-based explanations facilitate the transfer of prior knowledge into a new problem-solving context \cite{simon1979}. The inclusion of both teaching techniques has the potential to trigger sudden realisations of previously incomprehensible concepts that might lead to innovative learning outcomes. In contrast, the absence of these teaching mechanisms might hinder human learning of complex tasks and lead to more laborious problem-solving strategies. However, further investigations are needed to gain a better understanding of the degree to which sequential teaching curricula with machine-learned explanations facilitate novel human comprehension.
\section*{Declarations}

\begin{itemize}
\item Funding
\bigskip

This research was partially supported by TAILOR, a project funded by EU Horizon 2020 research and innovation programme under GA No. 952215. The third author acknowledges support from the UK’s EPSRC Human-Like Computing Network.
\bigskip

\item Conflict of interest/Competing interests 
\bigskip

The authors have no conflicts of interest to disclose.  
\bigskip

\item Ethics approval 
\bigskip

Ethics approval was not required. Included data and experiments were conducted by Imperial College London. Prior to carrying out the experiment, the authors had referred to the ethics approval guideline provided by Imperial College London (https://www.imperial.ac.uk/research-ethics-committee/ethics-approval-overview/). The authors had assessed that none of the five criteria applied to the planned experiment. Hence, the judgement was "No, we do not need to apply". The authors confirmed with the Research Governance and Integrity team at Imperial College London who accepted the authors' judgement.

\bigskip

\item Consent to participate
\bigskip

Informed consent for participation in the experiment was obtained from all individual participants included in the study.
\bigskip

\item Consent for publication
\bigskip

Informed consent for publication of data was obtained from all individual participants included in the study.
\bigskip

\item Availability of data and materials
\bigskip

Experimental material and data have been made available at

https://github.com/LAi1997/sequential-teaching.
\bigskip

\item Code availability 
\bigskip

Source code has been made available at

https://github.com/LAi1997/sequential-teaching.
\bigskip

\item Authors' contributions
\bigskip

Author 1 wrote all paper sections except Section \ref{sec:pre_test}. Author 2 wrote Section \ref{sec:pre_test}. Author 2, 3 and 4 provided feedback and corrections on all paper sections.  
\bigskip

\end{itemize}

\bibliographystyle{abbrv}
\bibliography{08-relatedwork,08-framework,08-cognitive,08-materials,08-conclusion}
\end{document}